\pdfoutput=1

\documentclass[11pt]{article}

\usepackage[final]{acl}

\usepackage{times}
\usepackage{latexsym}

\usepackage[T1]{fontenc}

\usepackage[utf8]{inputenc}

\usepackage{microtype}

\usepackage{inconsolata}

\usepackage{graphicx}

\usepackage{pifont}

\usepackage{placeins}
\usepackage{multirow}
\usepackage{arydshln}
\usepackage{graphicx}
\usepackage{paralist}
\usepackage{amssymb}
\usepackage{amsmath}
\usepackage{pifont}
\usepackage{xcolor}
\usepackage{bbm}
\usepackage{listings}
\usepackage{xspace}
\usepackage{algorithm}
\usepackage{algpseudocode}

\usepackage{cleveref}
\crefformat{section}{\S#2#1#3} 
\crefformat{subsection}{\S#2#1#3}
\crefformat{subsubsection}{\S#2#1#3}

\usepackage{pifont}

\title{Synchronous Faithfulness Monitoring for Trustworthy \\ Retrieval-Augmented Generation}

\author{
Di Wu, Jia-Chen Gu, Fan Yin, Nanyun Peng, Kai-Wei Chang \\
University of California, Los Angeles \\ 
\texttt{\{diwu,fanyin20,violetpeng,kwchang\}@cs.ucla.edu, gujc@ucla.edu}
}

\newcommand{\modelname}{\textsc{SynCheck}\xspace}
\newcommand{\searchalgname}{\textsc{FOD}\xspace}

\begin{document}
\maketitle

\begin{abstract}

Retrieval-augmented language models (RALMs) have shown strong performance and wide applicability in knowledge-intensive tasks. However, there are significant trustworthiness concerns as RALMs are prone to generating unfaithful outputs, including baseless information or contradictions with the retrieved context. This paper proposes \modelname, a lightweight monitor that leverages fine-grained decoding dynamics including sequence likelihood, uncertainty quantification, context influence, and semantic alignment to synchronously detect unfaithful sentences. By integrating efficiently measurable and complementary signals, \modelname enables accurate and immediate feedback and intervention, achieving 0.85 AUROC in detecting faithfulness errors across six long-form retrieval-augmented generation tasks, improving prior best method by 4\%. Leveraging \modelname, we further introduce \searchalgname, a faithfulness-oriented decoding algorithm guided by beam search for long-form retrieval-augmented generation. Empirical results demonstrate that \searchalgname outperforms traditional strategies such as abstention, reranking, or contrastive decoding significantly in terms of faithfulness, achieving over 10\% improvement across six datasets. 

\end{abstract}
\section{Introduction}

Retrieval-augmented language models (RALMs) synergize large language models (LLMs) with external knowledge sources such as Wikipedia, web search, or tool invocations \citep{lewis2020retrieval, pmlr-v119-guu20a, borgeaud2022improving,schick2024toolformer}. Recent studies show that directly providing free-formed retrieved evidence in the context of LLMs can correct inaccurate or outdated parametric knowledge \citep{ram-etal-2023-context, shi2023replug},
leading to strong performance on knowledge-intensive tasks such as open-domain question answering \citep{mallen-etal-2023-trust} and long-form generation \citep{asai2023self}.

Despite the promising performance, the \textit{trustworthiness} of RALMs' generation has become a concern: human evaluations reveal a substantial number of claims generated by RALMs contradicting with the provided context or cannot be grounded to any evidence \citep{wu2023ragtruth, wu2024well}. Such \textit{unfaithful} use of knowledge by LLMs renders it difficult to trust the output even if the knowledge source is proven trustworthy. Towards a fully faithful and transparent use of knowledge,  one line of work proposes \textit{post-hoc} attribution or revision \citep{gao-etal-2023-rarr, gao-etal-2023-enabling}, yet they are computationally expensive and could only be triggered after decoding. On the other hand, several \textit{synchronous} decoding interventions have been proposed for critiquing and correcting RALMs' outputs on-the-fly, such as dynamic retrieval \citep{jiang-etal-2023-active}, reranking with fine-tuned critique tokens \citep{asai2023self}, and contrastive decoding for amplifying the influence of the knowledge \citep{shi2023trusting}. However, these methods are mainly accuracy-oriented and it is unclear how well these signals are able to distinguish faithful samples from unfaithful ones. More importantly, these algorithms cannot provide any ways to guarantee or control the level of faithfulness of RALMs' final output.

\begin{figure*}[t!]
\centering
\includegraphics[width=\textwidth]{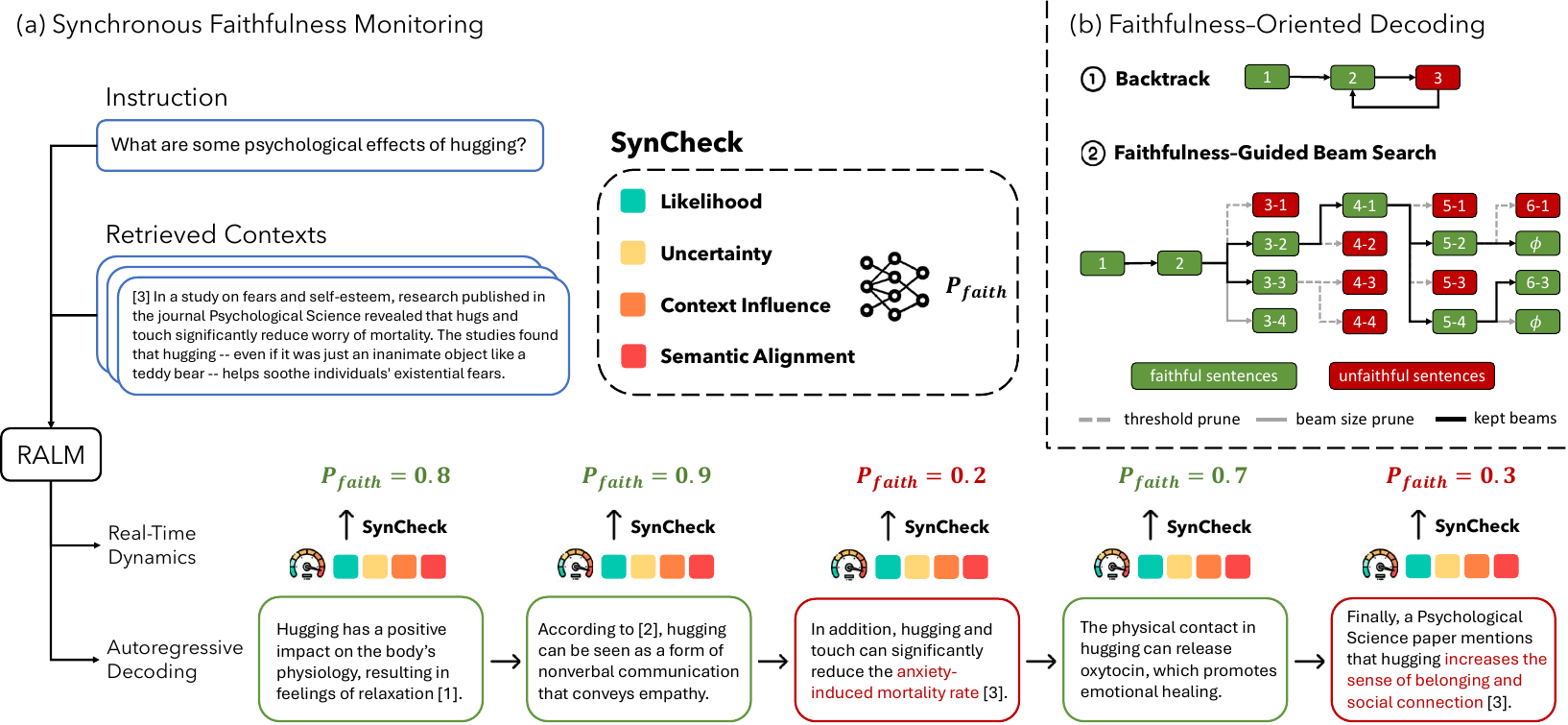}
\caption{(a) An illustration of \modelname, a fine-grained faithfulness checker for RALMs. \modelname aggregates real-time decoding dynamics to accurately judge whether a sentence is trustworthy or not. (b) Leveraging \modelname, we design a two-staged faithfulness-oriented decoding pipeline consisting of backtracking and beam search. In our algorithm, threshold prune takes place before beam size prune, and we use beam size 2 in the example.}
\label{main-framework}
\end{figure*}

To bridge these gaps, this paper undertakes a principled approach to faithfulness-oriented detection and decoding for long-form generation of RALMs. To start with, we compile a comprehensive benchmark to thoroughly evaluate faithfulness detectors at \textit{sentence-level}, with the tasks covering biography generation, question answering, summarization, and data-to-text. Surprisingly, existing quality control methods in RALM systems including likelihood-based filtering \citep{jiang-etal-2023-active} and instruction-tuned critique tokens \citep{asai2023self} exhibit serious deficiency, only achieving approximately 0.6 AUROC across all the tasks (\cref{results-monitoring}). 

In response, we propose \modelname, a simple, efficient, yet accurate faithfulness error detector. To accurately characterize different types of faithfulness errors, \modelname computes and monitors multiple complementary signals during RALM decoding: sequence likelihood, uncertainty quantification, context influence, and semantic alignment (Figure \ref{main-framework}a). The signals are efficiently measured \textit{synchronously} in the decoding process, and ensembled via a lightweight aggregator that imposes minimal overhead. Despite the efficient design, \modelname achieves a strong faithfulness judgment performance, scoring on average over 0.85 AUROC across six datasets and two LLMs, outperforming five traditional baselines by ranging from 4\% to 35\% (\cref{results-monitoring}). Furthermore, we find task-specific or model-specific training is not strictly required for training \modelname's aggregator, which significantly improves its usability (\cref{results-analysis}).

With \modelname, is it possible to design an intelligent and  efficient algorithm to improve the generation's faithfulness with a guarantee? Traditional methods struggle to strike a balance between informativeness and faithfulness. For instance, abstention \citep{kamath-etal-2020-selective, ren2022out, feng2024don} can be overly conservative, wasting the high quality part of the response. By contrast, contrastive decoding \citep{shi2023trusting} fails to enforce hard constraints to guarantee a basic level of faithfulness. We propose \textit{faithfulness-oriented decoding} (\searchalgname), a novel decoding algorithm that leverages \modelname to \textit{synchronously} monitor the generation faithfulness and guide the decoding process towards producing more faithful outputs.
Specifically, \searchalgname entails (1) backtracking at a low-quality sentence and (2) initiating a beam search that uses the faithfulness score to prune samples and guide the search direction (Figure \ref{main-framework}b). Experiments show that FOD significantly improves the generation's faithfulness over greedy search (12\%), abstention (10\%), reranking (13\%), and context-aware decoding (CAD, \citet{shi2023trusting}) (19\%) across six datasets. Compared to abstention, \searchalgname improves in both faithfulness and informativeness. Compared to CAD, FOD generates more faithful samples at the same number of sentences (\cref{results-intervention}). We will publicly release the benchmark and our code at \url{https://github.com/xiaowu0162/sync-ralm-faithfulness}.

\section{Related Work}

\paragraph{Context-Faithful LLMs}

Characterizing and improving the context faithfulness of LLMs have been an important research topic. \citet{longpre-etal-2021-entity} identify over-reliance of language models on their parametric knowledge when presented with contexts contradicting with it. \citet{xie2023adaptive} find that the persuasiveness of retrieved knowledge can be improved with an LM-centric knowledge verbalization process, yet LLMs still have inherent bias in choosing between conflicting pieces of knowledge. To improve the context faithfulness, common approaches include adapting the LLM to context-based generation \citep{shi2023context}, improving the context quality \citep{xu2023recomp}, improving the decoding method \citep{zhou-etal-2023-context,shi2023trusting}, and post-hoc detection or revision \citep{wu2023ragtruth, gao-etal-2023-rarr,gao-etal-2023-enabling}. Unfaithfulness to the context is also named as context-conflicting hallucinations \citep{zhang2023siren}. Notable related  works include detection with model features \citep{kadavath2022language, azaria2023internal, yin2024characterizing, jiang2024large}, decoding-stage interventions \citep{li2024inference, chuang2023dola, wang2024fine}, and abstention \citep{kamath-etal-2020-selective, ren2022out, feng2024don}. In this work, we show that traditional approaches adopted by RALM systems cannot accurately distinguish faithful generated sentences from unfaithful ones. By utilizing both model-centric features and sematic alignment, \modelname is able to characterize diverse types of unfaithfulness issues precisely, efficiently, and in a fine-grained manner.

\paragraph{Advanced Decoding for RALMs}
Recently, there has been a research interest in improving the decoding of black-box RALMs. \citet{trivedi-etal-2023-interleaving} and \citet{shao-etal-2023-enhancing} consider iterative retrieval for refining the generation. \citet{shi2023trusting} use contrastive decoding to emphasize the influence of the retrieved context. FLARE \citep{jiang-etal-2023-active} actively updates the context by setting a threshold on the likelihood of the least confident token in the generated sentence. Toolformer \citep{schick2024toolformer} learns to actively invoke tools at knowledge-intensive locations in generation. Self-RAG \citep{asai2023self} learns critique tokens to score and rerank hypotheses generated based on different retrieved documents. Compared to prior work, our decoding algorithm utilizes \modelname to give a fine-grained accurate guarantee of output quality, striking a better balance between faithfulness and informativeness of the output. 

\section{Approach}
\label{section-monitoring}

\subsection{Problem Formulation}
We consider retrieval augmented generation of free-form long responses. Let $\mathbf{x}$ be a sequence of input tokens encoding a question or an instruction. Let $\mathbf{c}$ denote a sequence of retrieved context tokens, which may include multiple free-form text chunks from multiple sources. Given the concatenation $[\mathbf{x};\mathbf{c}]$, 
an LLM $\mathcal{M}$ predicts a sequence of segments $(\mathbf{s}_1, ..., \mathbf{s}_m)$ as the response, where each segment consists of a sequence of tokens\footnote{In this paper, we treat one sentence as a segment in our experiments, but our framework is applicable to any segment granularity (i.e., multi-sentence or sub-sentence).}.

Then, the task of \textit{context faithfulness tracking} is defined as assigning a faithfulness label $\hat{y}_i\in\mathbb{R}$ to each newly generated segment $s_i$ based on $\mathbf{x}$, $\mathbf{c}$, and optionally the other segments. As many detection methods produce real-valued scores, we adopt AUROC as the evaluation metric.

\subsection{\modelname: Accurate Synchronous Faithfulness Monitoring for RALMs}
\label{detection-approach}

Different from factuality, faithfulness imposes a unique focus on checking whether an RALM comprehends the retrieved contexts and avoids deviating from the contexts. Such behavior is thus challenging to characterize using simple heuristics employed by previous literature such as the generative likelihood \citep{jiang-etal-2023-active}. Instead, we 
introduce \modelname, a fast and accurate feature-based detector for a range of untrustworthy generation scenarios on \textit{an ascending hierarchy}: (1) the parametric and context knowledge is insufficient for forming the response, (2) the model fails to utilize the context in its predictions, and (3) the model uses the context unfaithfully. Concretely, \modelname monitors four types of signals synchrnous to autoregressive decoding (Figure \ref{main-framework}a): 

\paragraph{\textbf{Likelihood}} Low likelihood outputs often indicate the presence of \textit{knowledge gaps}. When neither the parametric knowledge nor the retrieved information is sufficient, the model's response trivially bears faithfulness and trustworthiness issues. To detect this behavior, \modelname measures the minimum likelihood as well as the length-normalized likelihood across all tokens in each sentence $s_i$. 

\paragraph{\textbf{Uncertainty}} A high predictive uncertainty suggests an \textit{unconfident use of knowledge}. Whether the underlying cause is out-of-distribution questions, noisy retrieval, or a weak ability to incorporate the knowledge, the presence of high uncertainty strongly signals for a verification of the model's generation. \modelname monitors the averaged token-level entropy within $s_i$ as well as the local intrinsic dimension of the activation of intermediate layers, which we hypothesize to more precisely characterize the degree of the LLM unfaithfully mixing the retrieval context distribution with its parametric knowledge \citep{yin2024characterizing}. 

\paragraph{\textbf{Context Influence}} An important behavior previous work failed to capture is the \textit{over-dominance of parametric knowledge}. When an RALM heavily relies on its parametric knowledge, the generated information may deviate from the context and is thus largely non-attributable. Therefore, \modelname monitors two token-wise distributions: $P_{\mathcal{M}}({s_i | \mathbf{x}; \mathbf{c}; s_{1:i-1}})$ and $P_{\mathcal{M}}({s_i | \mathbf{x}; s_{1:i-1}})$, where $s_{1:i-1}$ is the sentences already generated by the RALM\footnote{Efficiency-wise, the second distribution indeed requires a separate forward pass. However, it could be calculated in parallel with the first distribution during token-level decoding.}. By contrasting the two distributions via token-level Kullback-Leibler divergence \citep{DBLP:journals/corr/abs-2306-01286}, we obtain informative indications of where $\mathbf{c}$ only has a weak influence on. 

\paragraph{\textbf{Semantic Alignment}} Even if the retrieved context exerts a high influence and the model produces a high confidence sentence, the output could still suffer from a \textit{misinterpretation of context}. When the model makes such a mistake, it is hard to detect and correct with model-centric features, To complement with the previous features, \modelname also runs a lightweight entailment checker \citep{zha-etal-2023-alignscore} to gauge the likelihood of each $s_i$ being semantically inconsistent with the retrieved $\mathbf{c}$. 

\paragraph{\modelname} For each $s_i$, we have collected a range of on-the-fly faithfulness signals. As they are designed to capture unique aspects, \modelname trains a light-weight aggregator to learn the task-specific decision boundary with a small labelled dataset\footnote{Empirically, the labelled set need not to be model-specific or task-specific, as we will explore in \cref{results-analysis}.}. We explore three hypothesis spaces: logistic regression, XGBoost \citep{chen2016xgboost}, and MLP. In the rest of the paper, they will be denoted as \textsc{SynCheck}\textsubscript{LR}, \textsc{SynCheck}\textsubscript{XGB}, and \textsc{SynCheck}\textsubscript{MLP} respectively. 

In \cref{appendix-feature-impl}, we document the formulation and implementation details regarding the decoding-time feature collection and the aggregator training.

\subsection{\searchalgname: Faithfuless-Oriented Decoding}
\label{section-intervention}

With the on-the-fly monitoring signals produced by \modelname, can we further design effective intervention approaches to improve the faithfulness of RALMs' output? Abstention, or selective prediction, is a straightforward application: after detecting potential quality issues, the system can refuse to generate any output \citep{kamath-etal-2020-selective, ren2022out, feng2024don}. However, the coarse-grained binary abstention decision wastes considerable faithful information generated by the model. To devise a more principled way to enhance the faithfulness of the output while increasing the information retained, we introduce \searchalgname, a \textit{faithfulness-oriented decoding} algorithm for RALMs. Concretely, \searchalgname integrates \modelname to guide the search direction synchronously, with two stages:

\begin{algorithm}[t!]
\caption{\small\textbf{\searchalgname: Faithfulness-Oriented Decoding}}
\small
\label{algo-fbs}
\begin{algorithmic}[1]
\Require \textsc{LLM}, \modelname, $\mathbf{x}$, $\mathbf{c}$, backtrack threshold $\tau_1$, sample pruning threshold $\tau_2$, sample size $S$, beam size $K$
\State // Stage 1: Greedy Search and Backtrack
\State \texttt{out} $\gets []$, $i \gets 0$
\State // next sentence $s_i$, faithfulness score $f_i$
\State $s_i, f_i \gets$ \modelname($\mathbf{x}$, $\mathbf{c}$, \textsc{LLM}($\mathbf{x}$; $\mathbf{c}$))
\While{$f_i \ge \tau_1$}
    \State \texttt{out} $\gets$ \texttt{out} + $[s_i]$, $i \gets i + 1$
    \State $s_i,f_i \gets$ \modelname($\mathbf{x}$, $\mathbf{c}$, \textsc{LLM}($\mathbf{x}$; $\mathbf{c}$; $s_{0:(i-1)}$))
\EndWhile

\\

\State // Stage 2: Faithfulness-Guided Beam Search
\State \texttt{beams} $\gets [$\texttt{out}$]$, \texttt{new\_beams} $\gets []$
\While{no beam has generated \texttt{[EOS]}}
    \For{$\mathcal{B}$\ in\ \texttt{beams}}
        \For{$j = 1, ..., \lceil S/K\rceil$}
            \State $s_i,f_i \gets$ \modelname($\mathbf{x}$, $\mathbf{c}$, \textsc{LLM}($\mathbf{x}$; $\mathbf{c}$; $\mathcal{B}$))
            \If {$f_i \ge \tau_2$}
                \State \texttt{new\_beams} $\gets$ $\mathcal{B} + [s_i]$
            \EndIf
        \EndFor
    \EndFor
    \If{\texttt{new\_beams}}
        \State \texttt{beams} $\gets$ $K$ most faithful beams in \texttt{new\_beams}
        \State $i \gets i + 1$, \texttt{\texttt{new\_beams}} $\gets []$
    \Else 
        \State \textbf{break}
    \EndIf
\EndWhile
\State \Return the most faithful beam in \texttt{beams}
\end{algorithmic}
\end{algorithm}

\begin{compactenum}
    \item Run greedy search until the first sentence $s_i$ with faithfulness score below a threshold $\tau_1$, which triggers the \textit{backtrack} operation.
    \item Starting from $s_{i-1}$, search in $K$ beams \textit{in parallel}. In each step, sample several continuations from each of the beams, directly pruning out samples that have faithfulness score below a threshold $\tau_2$. Finally, retain $K$ beams with the highest aggregated faithfulness score. 
\end{compactenum}

Figure \ref{main-framework}b presents an intuitive illustration of \searchalgname, and the full algorithm is presented in Algorithm \ref{algo-fbs}. We note that to calculate the faithfulness score of a partially decoded sequence, we always simply average across all the sentences in it. In addition, although we choose to halt the decoding when a single beam produces \texttt{[EOS]} as it gives empirically good performance, it is possible to continue the search until all the beams halt. 

\searchalgname's design optimizes the trade-off between faithfulness and informativeness. Compared to abstention, \searchalgname is often able to retain a faithful prefix that already contains substantial information. Moreover, compared to contrastive decoding methods that operate on token-level logit \citep{shi2023trusting} which does not guarantees on the final output's quality, \searchalgname provides a way to bound the minimum sentence-level faithfulness. Finally, the intermediate \modelname signals also make the decoding process more human interpretable, facilitating further downstream explanation or debugging.

\section{Experimental Setup}

\label{experimental-setup}

\subsection{Task and Dataset Collection}

We test \modelname and \searchalgname on a benchmark covering four common long-form RAG tasks: question answering (QA), summarization (Summ), data-to-text (Data2txt), and biography generation.

\paragraph{RAGTruth} For QA, Summ, and Data2txt, we use the questions and the retrieved contexts provided by RAGTruth \citep{wu2023ragtruth}, which are respectively sourced from MS MARCO \citep{bajaj2016ms}, CNN/Daily Mail \citep{see-etal-2017-get}, and Yelp Open Dataset \citep{yelpopendataset}. Notably, questions in RAGTruth are designed for \textit{long-form RAG}, which aligns well with our goal.

\paragraph{Biography} We use the factscore benchmark (\textbf{FS}) introduced in \citet{min-etal-2023-factscore}. To simulate situations where unfaithful generations are more likely to occur, we construct two new datasets named famous-100 (\textbf{F-100)} and famous-100-anti (\textbf{F-100-anti}) where a model is asked to write a biography for 100 famous entities. The context for F-100 are retrievd from wikipedia, and for F-100-anti we create contexts from the evidence retrieved for another entity through entity substitution. By querying popular and salient parametric knowledge, F-100 and F-100-anti creates challenging scenarios for resisting to generate baseless/conflicting information.

\paragraph{Split} We follow the train-test split in RAGTruth for QA, Summ, and Data2txt. FS, F-100, and F-100-anti only have a single test split. For \modelname training, the respective train sets are used for QA, Summ, and Data2txt. The models for F-100 and F-100-anti are trained on FS, and the model for FS is trained on F-100. We further document the dataset construction, split details, as well as the basic statistics in \cref{appendix-benchmark-details}.

\subsection{Context Faithfulness Tracking}

\paragraph{Output Collection} We mainly test on Llama 2 7B Chat \citep{touvron2023llama} and Mistral 7B Instruct \citep{jiang2023mistral}\footnote{We use the model distributed at \url{https://huggingface.co/mistralai/Mistral-7B-Instruct-v0.1}.}. For FS, F-100, and F-100-anti, we collect the outputs via greedy decoding. For QA, Summ, and Data2txt, we directly leverage the outputs provided by RAGTruth, which were sampled using various temperatures. 

\paragraph{Label construction} We use NLTK\footnote{\url{https://www.nltk.org/api/nltk.tokenize.PunktSentenceTokenizer.html}} to decompose the output into sentences, and assign each sentence with a faithfulness label. For QA, Summ, and Data2txt, we use the human-annotated baseless spans and conflict spans \citep{wu2023ragtruth} as the unfaithful spans. For FS, F-100, and F-100-anti, we use a pre-trained propositionizer \citep{chen2023dense} to decompose the outputs into decontextualized propositions and then use an AutoAIS model \citep{honovich-etal-2022-true-evaluating} to judge the faithfulness of each proposition. Finally, a lexical matching algorithm is used to map the span/proposition level faithfulness labels into sentence-level labels. We provide the detailed algorithm in \cref{appendix-trustworthy-label-generation}.

\setlength{\tabcolsep}{3.5pt}
\begin{table*}[t!]
 \centering
 \resizebox{0.85\linewidth}{!}{
 \begin{tabular}{c | c c c | c c c | c }
 \hline
 \multirow{2}{*}{\textbf{Method}} & \multicolumn{3}{c|}{\textbf{RAGTruth}} & \multicolumn{3}{c|}{\textbf{Biography Generation}} &  \multirow{2}{*}{\textbf{Average}}  \\
& \textbf{QA} & \textbf{Summ} & \textbf{Data2txt} & \textbf{FS} & \textbf{F-100} & \textbf{F-100-anti} \\
 \hline
 \multicolumn{8}{c}{\textbf{\texttt{Llama 2 7B Chat}}} \\
 \hdashline
 \textsc{SpanExtract}\textsubscript{GPT-4-Turbo}\textsuperscript{\ding{67}} & 0.705 & 0.773 & 0.794 & 0.630 & 0.548 & 0.506 & 0.659 \\
 \textsc{SpanExtract}\textsubscript{Llama-2-13B}\textsuperscript{\ding{67}} & 0.734 & 0.688 & \textbf{0.842} & 0.552 & 0.660 & 0.657 & 0.689 \\
 \textsc{AlignScore}\textsuperscript{\ding{67}} & 0.772 & 0.768 & 0.679 & \textbf{0.791} & 0.897 & 0.886 & 0.799 \\
\textsc{MiniCheck}\textsuperscript{\ding{67}} & 0.788 & 0.778 & 0.813 & 0.736 & 0.833 & 0.869 & 0.803 \\
 \textsc{CriticTok}\textsuperscript{\ding{95}} & 0.506 & 0.534 & 0.604 & 0.565 & 0.767 & 0.784 & 0.627 \\
 \textsc{FLARE}\textsuperscript{\ding{95}} & 0.631 & 0.527 & 0.532 & 0.649 & 0.694 & 0.677 & 0.618 \\
\textsc{SynCheck}\textsubscript{LR}\textsuperscript{\ding{95}} & 0.812 & 0.753 & 0.775 & 0.771 & 0.908 & 0.893 & 0.819 \\
\textsc{SynCheck}\textsubscript{XGB}\textsuperscript{\ding{95}} & 0.803 & 0.730 & 0.752 & 0.752 & 0.901 & 0.866 & 0.801 \\
\textsc{SynCheck}\textsubscript{MLP}\textsuperscript{\ding{95}} & \ \ \textbf{0.833}$_3$ & \ \ \textbf{0.787}$_3$ & \ \ 0.785$_2$ & \ \ 0.764$_4$ & \ \ \textbf{0.918}$_3$ & \ \ \textbf{0.896}$_2$ & \textbf{0.831} \\
 \hline
 \multicolumn{8}{c}{\textbf{\texttt{Mistral 7B Instruct}}} \\
 \hdashline
 \textsc{SpanExtract}\textsubscript{GPT-4-Turbo}\textsuperscript{\ding{67}} & 0.775 & 0.830 & 0.827 & 0.529 & 0.648 & 0.431 & 0.673 \\
 \textsc{SpanExtract}\textsubscript{Llama-2-13B}\textsuperscript{\ding{67}} & 0.641 & 0.716 & \textbf{0.872} & 0.555 & 0.660 & 0.704 & 0.691 \\
 \textsc{AlignScore}\textsuperscript{\ding{67}} & 0.875 & 0.829 & 0.731 & \textbf{0.804} & 0.956 & 0.871 & 0.844 \\
 \textsc{MiniCheck}\textsuperscript{\ding{67}} & 0.824 & \textbf{0.874} & 0.832 & 0.761 & 0.852 & 0.880 & 0.837 \\
 \textsc{CriticTok}\textsuperscript{\ding{95}} & 0.588 & 0.578 & 0.529 & 0.564 & 0.870 & 0.838 & 0.661 \\
 \textsc{FLARE}\textsuperscript{\ding{95}} & 0.539 & 0.603 & 0.463 & 0.690 & 0.798 & 0.638 & 0.622 \\
 \textsc{SynCheck}\textsubscript{LR}\textsuperscript{\ding{95}} & \textbf{0.895} & 0.785 & 0.731 & 0.765 & \textbf{0.975} & 0.871 & 0.837 \\
 \textsc{SynCheck}\textsubscript{XGB}\textsuperscript{\ding{95}} & 0.877 & 0.750 & 0.819 & 0.769 & 0.865 & 0.867 & 0.825 \\
 \textsc{SynCheck}\textsubscript{MLP}\textsuperscript{\ding{95}} & \ \ 0.887$_8$ & \ \ 0.829$_0$ & \ \ 0.856$_2$ & \ \ 0.773$_2$ & \ \ 0.972$_2$ & \ \ \textbf{0.883}$_3$ & \textbf{0.867} \\
 \hline
\end{tabular}
 }
 \caption{AUROC results of all context faithfulness tracking methods. We use superscripts to differentiate between external lexical detectors (\ding{67}) and methods using RALM-centric features (\ding{95}). We use subscript to denote the standard deviation across three runs. For instance, 0.887$_8$ means a mean 0.887 and a standard deviation 0.008. \textsc{SynCheck}\textsubscript{MLP} achieves the strongest performance averaged across six tasks for both LLMs experimented. }
 \label{tab:detection_acc_main}
\end{table*}

\paragraph{Baselines}

We compare \modelname with the following faithfulness checking baselines: 

\begin{itemize}
    \item \textsc{SpanExtract}. \citet{wu2023ragtruth} proposes to instruct an evaluator LLM to directly predict the spans from the output corresponding to the unfaithful statements. We test GPT-4-Turbo and the fine-tuned Llama 2 13B model provided by \citet{wu2023ragtruth}.
    \item \textsc{CriticTok}. Self-RAG \citep{asai2023self} is an LLM instruction-tuned for critiquing RAG outputs. We leverage the model's critic token \texttt{[IsSup]} and report the score as the probability of \texttt{[fully supported]} divided by the sum of the probability of \texttt{[partially supported]} and \texttt{[no support]}, with each $s_i$ as the generated segment to critique.
    \item \textsc{FLARE}. Following \citet{jiang-etal-2023-active}, we use the minimum likelihood across all the tokens in the generated sentence $s_i$.
    \item Lexical Alignment Models. We test \textsc{AlignScore} \citep{zha-etal-2023-alignscore} and \textsc{MiniCheck} \citep{tang2024minicheck}, both of which calculate a semantic alignment score between $\mathbf{C}$ and $s_i$.
\end{itemize}

In \cref{appendix-baseline-implementation}, we provide the implementation details of the baselines, including the prompt format for \textsc{SpanExtract} and \textsc{CriticTok}.

\subsection{Faithfulness Intervention}
\label{experimental-setup-intervention}

We use the same set of datasets and models to evaluate the proposed decoding method \searchalgname. 

\paragraph{Evaluation Metric} We report two \textit{response-level} metrics: faithfulness and informativeness. Inspired by \citet{min-etal-2023-factscore}, faithfulness is designed as proposition-level contextual consistency. The model proposed in \citet{chen2023dense} is used to decompose the response into propositions, and we use \texttt{retrieval+llama+npm} method \citep{min-etal-2023-factscore} to factcheck each proposition directly with $\mathbf{C}$ as the context. Faithfulness is reported as the proportion of faithful propositions within all propositions. For informativeness, we report the number of propositions in the response. For the abstained or empty responses, we exclude them from faithfulness evaluation but assign 0 as the informativeness score.

\setlength{\tabcolsep}{3.5pt}
\begin{table*}[ht!]
 \centering
 \resizebox{0.95\linewidth}{!}{
 \begin{tabular}{c | c c | c c | c c | c c | c c | c c | c c }
 \hline
 \multirow{2}{*}{\textbf{Method}} & \multicolumn{2}{c|}{\textbf{QA}} & \multicolumn{2}{c|}{\textbf{Summ}} & \multicolumn{2}{c|}{\textbf{Data2txt}} & \multicolumn{2}{c|}{\textbf{FS}} & \multicolumn{2}{c|}{\textbf{F-100}} & \multicolumn{2}{c|}{\textbf{F-100-anti}} & \multicolumn{2}{c}{\textbf{Average}}\\
 & Faith. & Info. & Faith. & Info. & Faith. & Info. & Faith. & Info. & Faith. & Info. & Faith. & Info. & Faith. & Info. \\
 \hline
 \multicolumn{15}{c}{\textbf{\texttt{Llama 2 7B Chat}}} \\
 \hdashline
 Greedy & 0.628 & 10.9 & 0.716 & 9.4 & 0.440 & 13.3 & 0.489 & 8.0 & 0.766 & 9.5 & 0.603 & 7.6 & 0.607 & 9.8 \\
 CAD & 0.710 & 10.4 & 0.755 & 9.5 & 0.468 & 14.0 & 0.278 & 7.4 & 0.611 & 22.4 & 0.476 & 19.2 & 0.549 & 13.8 \\
 Abstention\textsuperscript{\ding{95}} & 0.639 & 3.5 & 0.712 & 9.0 & 0.422 & 8.2 & 0.509 & 2.8 & 0.809 & 2.3 & 0.584 & 1.1 & 0.613 & 3.7 \\ 
 Reranking\textsuperscript{\ding{95}} & 0.745 & 9.3 & 0.757 & 9.8 & 0.474 & 13.4 & 0.466 & 10.4 & 0.621 & 10.0 & 0.563 & 12.9 & 0.604 & 11.0  \\
 \searchalgname (BT)\textsuperscript{\ding{95}} & 0.737 & 4.0 & 0.744 & 8.4 & \textbf{0.556} & 7.3 & \textbf{0.532} & 4.5 & 0.843 & 4.5 & 0.668 & 3.0 & 0.680 & 5.3 \\
 \searchalgname (Full)\textsuperscript{\ding{95}} & \textbf{0.768} & 4.3 & \textbf{0.770} & 9.5 & 0.529 & 10.4 & 0.507 & 5.1 & \textbf{0.870} & 4.8 & \textbf{0.735} & 3.5 & \textbf{0.697} & 6.3 \\
 \hline
 \multicolumn{15}{c}{\textbf{\texttt{Mistral 7B Instruct}}} \\
 \hdashline
 Greedy & 0.725 & 7.4 & 0.737 & 10.2 & 0.431 & 14.4 & 0.482 & 11.2 & 0.741 & 7.9 & 0.598 & 7.2 & 0.619 & 9.7 \\
 CAD & 0.759 & 7.1 & \textbf{0.799} & 11.6 & 0.378 & 17.9 & 0.389 & 11.9 & 0.698 & 13.7 & 0.539 & 13.9 & 0.594 & 12.7 \\
 Abstention\textsuperscript{\ding{95}} & 0.701 & 0.9 & 0.757 & 7.7 & 0.441 & 11.9 & 0.456 & 2.8 & 0.737 & 3.1 & 0.706 & 2.3 & 0.633 & 4.8 \\
 Reranking\textsuperscript{\ding{95}} & 0.737 & 6.2 & 0.802 & 11.2 & 0.457 & 14.9 & 0.395 & 12.5 & 0.652 & 7.5 & 0.552 & 11.3 & 0.599 & 10.6 \\
  \searchalgname (BT)\textsuperscript{\ding{95}} & 0.781 & 1.3 & 0.790 & 6.3 & \textbf{0.603} & 4.7 & \textbf{0.510} & 8.2 & 0.756 & 6.2 & 0.692 & 4.5 & \textbf{0.688} & 5.2\\
 \searchalgname (Full)\textsuperscript{\ding{95}} & \textbf{0.846} & 4.0 & 0.796 & 10.9 & 0.440 & 13.6 & 0.439 & 8.5 & \textbf{0.769} & 6.6 & \textbf{0.716} & 4.9 & 0.668 & 8.1 \\
 \hline
 \end{tabular}
 }
 \caption{Faithfulness-Informativeness evaluation results of faithfulness intervention methods. BT means backtracking only, without the following beam search. For all the results, we use a threshold 0.7 for abstention, and $\tau_1$=0.7, $\tau_2$=0.85 for our method, which we find generally work well. In addition, we use $K=2$ and $S=6$ for \searchalgname (Full) and sample size 6 for reranking. We use \ding{95} to mark the decoding methods that leverage \modelname.} 
 \label{tab:intervention_main_results}
\end{table*}

\begin{figure*}[t!]
\centering
\includegraphics[width=0.93\textwidth]{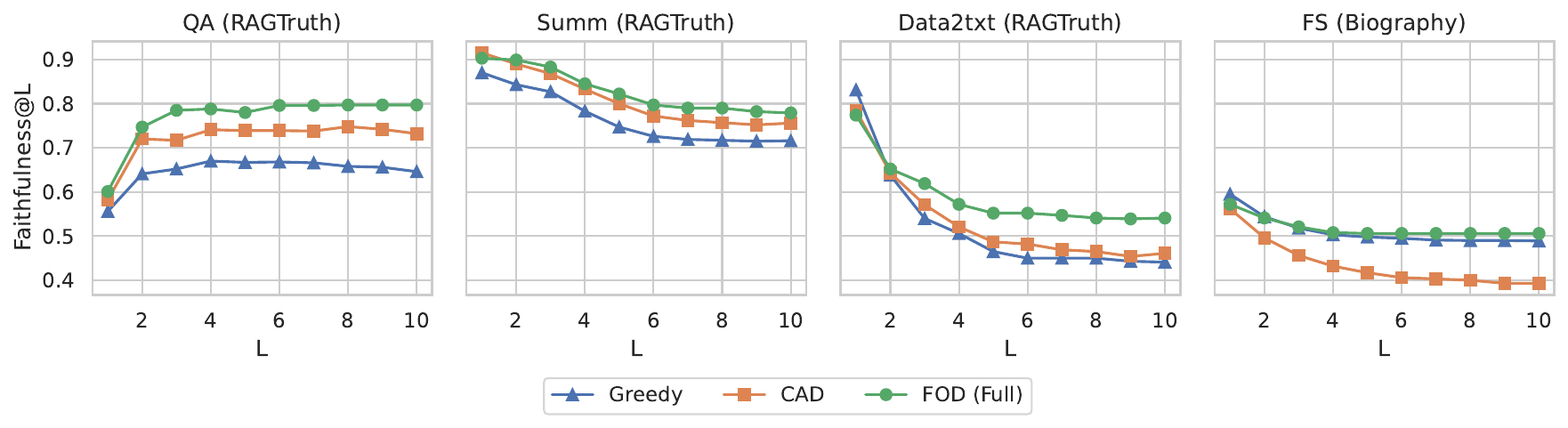}
\caption{Faithfulness score on Llama 2 7B chat with output truncated to the first $L$ sentences.}
\label{faithfulness-at-k-llama}
\end{figure*}

\paragraph{Baseline} We compare with (1) \textbf{abstention} - refusing to output when the faithfulness score predicted by \textsc{SynCheck}\textsubscript{MLP} for any of the output sentence below a certain threshold; (2) \textbf{reranking} - freely sampling the same number of responses as \searchalgname would and directly return the sample with the best averaged faithfulness score from \modelname; and (3) \textbf{CAD} \citep{shi2023trusting}, a contrastive decoding method amplifying the influence from the context.

\section{Results}

\subsection{Context Faithfulness Tracking}
\label{results-monitoring}

In Table \ref{tab:detection_acc_main}, we report the AUROC of all the context faithfulness tracking approaches we consider. To begin with, we find that traditional methods adopted by RALMs systems, such as \textsc{CriticTok} and \textsc{FLARE}, only provide limited accuracy in identifying whether a single sentence is faithful or not. Moreover, \textsc{SpanExtract} and \textsc{AlignScore} exhibit weak generalization performance across tasks. \textsc{SpanExtract} performs poorly on Biography Generation tasks, which it has not been optimized on. \textsc{AlignScore} achieves decent performance on QA, Summ, and Biography as these tasks are similar to its extensive pre-training data \citep{zha-etal-2023-alignscore}. However, AlignScore fails to generalize to Data2txt. while requires fine-grained checking of details such as locations, ratings, and numeric information.

Among all the methods, \textsc{SynCheck}\textsubscript{MLP} achieves the strongest performance averaged across all the tasks, with an average 0.831 AUROC for Llama 2 7B Chat and 0.867 for Mistral 7B Instruct. Notably, despite \modelname leverages \textsc{AlignScore} for semantic alignment, it is able to outperform \textsc{AlignScore} on most of the tasks for both models. In \cref{results-analysis}, we provide further insights on feature importance as well as cross-task and cross-model generalization of \modelname.  Finally, among the hypothesis spaces we explored, we surprisingly find that simple logistic regression can achieve a very strong detection performance. This further highlights the value of the ensembled signals themselves, which enables high performance with simple and lightweight aggregators.

\subsection{Intervention for Trustworthiness}
\label{results-intervention}

Next, we study leveraging \textsc{SynCheck}\textsubscript{MLP} to improve the trustworthiness of the model output.

\paragraph{Faithfulness vs. Informativeness} In Table \ref{tab:intervention_main_results}, we evaluate the faithfulness-informativeness trade-off of different decoding strategies. Compared to the greedy search, \searchalgname (BT) can already greatly improve the faithfulness. Compared to abstention, \searchalgname (BT) improves \textit{both} faithfulness and informativeness. For Summ, Data2txt, and FS, the response after backtracking still bears a number of propositions, indicated by the high informativeness. Through \searchalgname (Full), the informativeness is further boosted without significantly impacting faithfulness. Notably, the hyperparameters for \searchalgname ($\tau_1$, $\tau_2$, $K$, and $S$) are kept the same across all models and tasks, indicating its generalizability. Although CAD improves the faithfulness for QA, Summ, and Data2txt, it is ineffective for biography generation tasks. By comparison, \searchalgname still consistently outperforms CAD for all of the six tasks and two models, except Summ for Mistral where the two algorithms have similar performance. Finally, compared to sampling and post-hoc reranking \searchalgname significantly improve the faithfulness, indicating the necessity of performing on-the-fly monitoring and pruning of already unfaithful samples. 

\paragraph{Faithfulness@L} To check whether \searchalgname truly provides a better faithfulness-informativeness trade-off, we further compare it with CAD in a more controlled setting. In Figure \ref{faithfulness-at-k-llama}, we present the faithfulness of their predictions truncated to $L$ sentences. \searchalgname consistently improves over greedy search and CAD across all the tasks, further confirming the effectiveness of the algorithm.

\subsection{Analysis}
\label{results-analysis}

In this section, we provide further analyses on \modelname including feature ablation as well as cross-task/model faithfulness tracking.

\begin{figure}[t!]
\centering
\includegraphics[width=\linewidth]{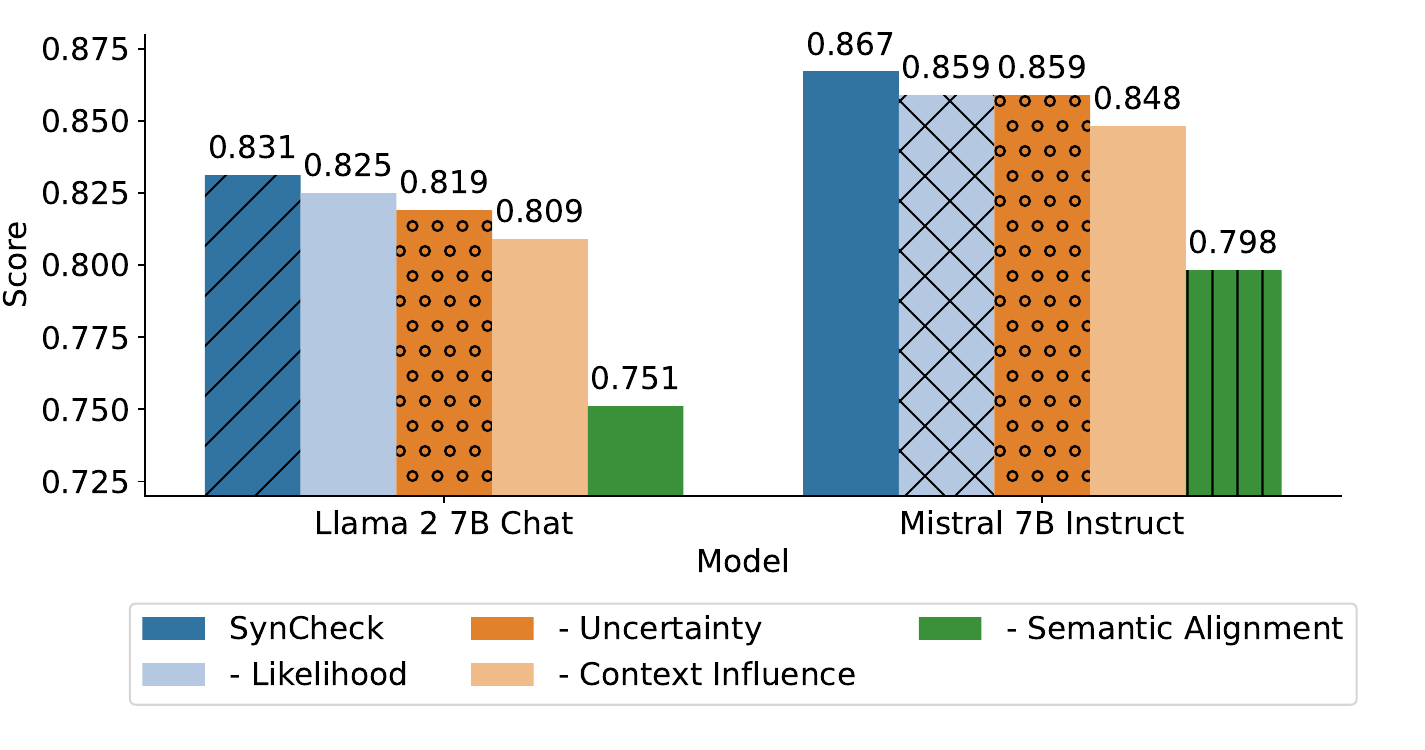}
\caption{Feature ablation study with \textsc{SynCheck}\textsubscript{MLP} used as the detection model. We report the AUROC averaged across three runs with different random seeds.}
\label{detector_feature_ablation}
\end{figure}

\paragraph{Ablation study} We conduct ablations on the features used by \modelname, and present the results in Figure \ref{detector_feature_ablation}. Overall, we observe that removing each type of the proposed feature can harm the detection performance for both of the models studied. Although semantic alignment is the most important feature, it is also necessary to combine it with other dimensions to achieve the state-of-the-art performance. Context influence is the second important dimension, causing 0.02 to 0.03 absolute drop in AUROC when removed, suggesting its necessity and the unique value in discerning segments that are unfaithful to the context. 

\begin{figure}[t!]
\centering
\includegraphics[width=0.9\linewidth]{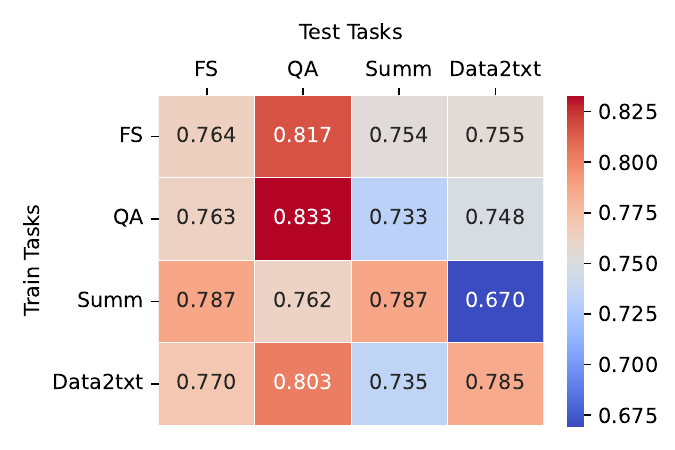}
\caption{Performance of \textsc{SynCheck}\textsubscript{MLP} on different train-test task pairs using Llama 2 7B Chat. We report the AUROC averaged across three runs.}
\label{cross_task_detection}
\end{figure}

\paragraph{Cross-Task Faithfulness Tracking} So far, we have assumed that task-specific data is required to train \modelname. In Figure \ref{cross_task_detection}, we investigate the AUROC of \textsc{SynCheck}\textsubscript{MLP} trained on tasks other than the tested task. Overall, we find that detectors trained on one task can often transfer decently to another task. Specifically, most of the tasks can transfer well to detect unfaithful generations on QA. In addition, detectors trained on FS or Data2txt transfer well to most of the other tasks. Crucially, this result relaxes the dependence on task-specific data, leading to a more generalizable and data-efficient trustworthiness detection approach.

\paragraph{Cross-Model Faithfulness Tracking} Finally, we explore using Llama 2 7B Chat as a \textit{surrogate model} \citep{shrivastava2023llamas} to perform faithfulness tracking for outputs generated by other models. As shown in Table \ref{tab:cross_model_detection}, \modelname trained on the surrogate model achieves a high performance on judging the faithfulness of other models' outputs. We hypothesize that although the samples may have a low likelihood to the surrogate model, other dimensions such as context influence and uncertainty still exhibit a clear boundary between faithful and unfaithful samples, making \modelname applicable without access of the checked model or task-specific trustworthiness labels. 

\setlength{\tabcolsep}{3.5pt}
\begin{table}[t!]
  \centering
  \resizebox{\linewidth}{!}{
  \begin{tabular}{l | c c c }
    \hline
    & \texttt{Llama (13B)} & \texttt{Llama (70B)} & \texttt{Mistral (7B)} \\
    \hline
    QA & 0.854 & 0.853 & 0.870 \\
    Summ & 0.811 & 0.739 & 0.821 \\
    Data2txt & 0.718 & 0.706 & 0.779 \\
    Bio & 0.782 & 0.759 & 0.777 \\
    F-100 & 0.897 & 0.908 & 0.977 \\
    F-100-anti & 0.866 & 0.865 & 0.906 \\
    \hdashline
    Average & 0.821 & 0.805 & 0.855 \\    
    \hline
  \end{tabular}
  }
  \caption{Faithfulness tracking with \textsc{SynCheck}\textsubscript{MLP} and Llama 2 7B Chat as the surrogate model.  \texttt{Llama} =  \texttt{Llama 2 Chat},  \texttt{Mistral} =  \texttt{Mistral Instruct}.}
  \label{tab:cross_model_detection}
  
\end{table}

We further provide a comparison between \searchalgname and single-feature reranking baselines in \cref{appendix-more-rerank-baselines}, a study for the hyperparameters $S$ and $K$ of \searchalgname in \cref{appendix-search-hyperparam}, as well as several qualitative examples in \cref{appendix-qualitative-study}.

\section{Conclusion}

In this paper, we introduced \modelname, a real-time lightweight detector designed to enhance the faithfulness of RALMs in long-form generation by leveraging fine-grained decoding dynamics to detect unfaithful generated segments. Our empirical results demonstrate that \modelname significantly outperforms existing faithfulness detection methods, achieving high AUROC scores across various tasks. Additionally, we proposed Faithfulness-Oriented Decoding, which improves the faithfulness and informativeness of RALM outputs compared to traditional faithfulness intervention methods. This work underscores the importance of real-time monitoring and targeted interventions in advancing RALMs as reliable tools for knowledge-intensive generation, paving the way for more trustworthy and interpretable RALM systems.

\section*{Limitations}

Despite the effectiveness of \modelname and \searchalgname, several limitations remain:

\begin{compactenum}
    \item \textbf{Latency}. Despite the \modelname operates logically concurrently with the decoding running it in beam search incurs extra latency. Specifically, in each step, $S$ sentences are sampled and have their faithfulness checked. With $S=1$, the latency cost of \searchalgname is exactly the same as CAD, since both of which incur an extra forward pass per token for calculating context influence. With $S$ GPUs, which is usually not a large number, \searchalgname achieves the same latency as CAD.
    \item \textbf{Segment Granularity}. In the paper, we mainly use sentences as the segment granularity, which is consistent with prior work such as \citep{jiang-etal-2023-active}. Future work could further confirm and improve the performance of \searchalgname with segments of different granularity such as sub-sentence or multi-sentence.
    \item \textbf{Diverse Real-World Tasks}. Finally, although we have performed evaluation on four representative tasks, further deploying the system to improve the quality of RAG on diverse real-world tasks in the wild is an exciting next step. In addition, it is also a promising investigation to employ \modelname as a plug-and-play faithfulness monitor with RAG pipelines that feature larger models as the generation model.
\end{compactenum}

\section*{Ethics Statement}

In this paper, we mainly investigate detecting and correcting the unfaithfulness in the outputs of RALMs. However, we are aware that faithfulness does not entail robustness, that a faithful RALM could be misled by biased resources and produce socially harmful claims. To build trustworthy RALM systems, we believe both the LLM and the retriever need the additional ability to critically judge the quality of the retrieved information before blindly following them. Finally, we are also aware that our approach improves the performance at the cost of additional computation. We are committed to further improving efficiency of our algorithm while maintaining the performance. 

We access the officially released datasets for the RAGTruth datasets and the FS entities, both of which are released under the MIT license. We use these datasets as-is without any additional preprocessing process. As OpenAI models are involved in curating the F-100 and F-100-anti dataset, and a range of models are involved in creating the faithfulness tracking data, our code and datasets will be released with MIT license with a research-only use permission. In addition, we will not re-distribute the RAGTruth data but will instead redirect to the version distributed by the original authors.

\section*{Acknowledgments}
This research was supported in part by Taboola, CISCO, and NSF \#2331966. We thank the Taboola team for the helpful discussion. We also thank I-Hung Hsu, Po-Nien Kung, Tanmay Parekh, as well as other members of the UCLA-NLP group and anonymous reviewers for their valuable feedback.

\bibliography{anthology,custom}

\begin{thebibliography}{46}
\providecommand{\natexlab}[1]{#1}

\bibitem[{Asai et~al.(2024)Asai, Wu, Wang, Sil, and Hajishirzi}]{asai2023self}
Akari Asai, Zeqiu Wu, Yizhong Wang, Avirup Sil, and Hannaneh Hajishirzi. 2024.
\newblock \href {https://openreview.net/forum?id=hSyW5go0v8} {Self-{RAG}: Learning to retrieve, generate, and critique through self-reflection}.
\newblock In \emph{The Twelfth International Conference on Learning Representations}.

\bibitem[{Azaria and Mitchell(2023)}]{azaria2023internal}
Amos Azaria and Tom Mitchell. 2023.
\newblock \href {https://arxiv.org/abs/2304.13734} {The internal state of an llm knows when its lying}.
\newblock \emph{ArXiv preprint}, abs/2304.13734.

\bibitem[{Bajaj et~al.(2016)Bajaj, Campos, Craswell, Deng, Gao, Liu, Majumder, McNamara, Mitra, Nguyen et~al.}]{bajaj2016ms}
Payal Bajaj, Daniel Campos, Nick Craswell, Li~Deng, Jianfeng Gao, Xiaodong Liu, Rangan Majumder, Andrew McNamara, Bhaskar Mitra, Tri Nguyen, et~al. 2016.
\newblock \href {https://arxiv.org/abs/1611.09268} {Ms marco: A human generated machine reading comprehension dataset}.
\newblock \emph{ArXiv preprint}, abs/1611.09268.

\bibitem[{Borgeaud et~al.(2022)Borgeaud, Mensch, Hoffmann, Cai, Rutherford, Millican, van~den Driessche, Lespiau, Damoc, Clark, de~Las~Casas, Guy, Menick, Ring, Hennigan, Huang, Maggiore, Jones, Cassirer, Brock, Paganini, Irving, Vinyals, Osindero, Simonyan, Rae, Elsen, and Sifre}]{borgeaud2022improving}
Sebastian Borgeaud, Arthur Mensch, Jordan Hoffmann, Trevor Cai, Eliza Rutherford, Katie Millican, George van~den Driessche, Jean{-}Baptiste Lespiau, Bogdan Damoc, Aidan Clark, Diego de~Las~Casas, Aurelia Guy, Jacob Menick, Roman Ring, Tom Hennigan, Saffron Huang, Loren Maggiore, Chris Jones, Albin Cassirer, Andy Brock, Michela Paganini, Geoffrey Irving, Oriol Vinyals, Simon Osindero, Karen Simonyan, Jack~W. Rae, Erich Elsen, and Laurent Sifre. 2022.
\newblock \href {https://proceedings.mlr.press/v162/borgeaud22a.html} {Improving language models by retrieving from trillions of tokens}.
\newblock In \emph{International Conference on Machine Learning, {ICML} 2022, 17-23 July 2022, Baltimore, Maryland, {USA}}, volume 162 of \emph{Proceedings of Machine Learning Research}, pages 2206--2240. {PMLR}.

\bibitem[{Chang et~al.(2023)Chang, Reitter, Aksitov, and Sung}]{DBLP:journals/corr/abs-2306-01286}
Chung{-}Ching Chang, David Reitter, Renat Aksitov, and Yun{-}Hsuan Sung. 2023.
\newblock \href {https://doi.org/10.48550/ARXIV.2306.01286} {Kl-divergence guided temperature sampling}.
\newblock \emph{CoRR}, abs/2306.01286.

\bibitem[{Chen and Guestrin(2016)}]{chen2016xgboost}
Tianqi Chen and Carlos Guestrin. 2016.
\newblock \href {https://doi.org/10.1145/2939672.2939785} {Xgboost: {A} scalable tree boosting system}.
\newblock In \emph{Proceedings of the 22nd {ACM} {SIGKDD} International Conference on Knowledge Discovery and Data Mining, San Francisco, CA, USA, August 13-17, 2016}, pages 785--794. {ACM}.

\bibitem[{Chen et~al.(2023)Chen, Wang, Chen, Yu, Ma, Zhao, Yu, and Zhang}]{chen2023dense}
Tong Chen, Hongwei Wang, Sihao Chen, Wenhao Yu, Kaixin Ma, Xinran Zhao, Dong Yu, and Hongming Zhang. 2023.
\newblock \href {https://arxiv.org/abs/2312.06648} {Dense x retrieval: What retrieval granularity should we use?}
\newblock \emph{ArXiv preprint}, abs/2312.06648.

\bibitem[{Chuang et~al.(2024)Chuang, Xie, Luo, Kim, Glass, and He}]{chuang2023dola}
Yung-Sung Chuang, Yujia Xie, Hongyin Luo, Yoon Kim, James~R Glass, and Pengcheng He. 2024.
\newblock \href {https://openreview.net/forum?id=Th6NyL07na} {Dola: Decoding by contrasting layers improves factuality in large language models}.

\bibitem[{Feng et~al.(2024)Feng, Shi, Wang, Ding, Balachandran, and Tsvetkov}]{feng2024don}
Shangbin Feng, Weijia Shi, Yike Wang, Wenxuan Ding, Vidhisha Balachandran, and Yulia Tsvetkov. 2024.
\newblock \href {https://doi.org/10.18653/v1/2024.acl-long.786} {Don{'}t hallucinate, abstain: Identifying {LLM} knowledge gaps via multi-{LLM} collaboration}.
\newblock In \emph{Proceedings of the 62nd Annual Meeting of the Association for Computational Linguistics (Volume 1: Long Papers)}, pages 14664--14690, Bangkok, Thailand. Association for Computational Linguistics.

\bibitem[{Gao et~al.(2023{\natexlab{a}})Gao, Dai, Pasupat, Chen, Chaganty, Fan, Zhao, Lao, Lee, Juan, and Guu}]{gao-etal-2023-rarr}
Luyu Gao, Zhuyun Dai, Panupong Pasupat, Anthony Chen, Arun~Tejasvi Chaganty, Yicheng Fan, Vincent Zhao, Ni~Lao, Hongrae Lee, Da-Cheng Juan, and Kelvin Guu. 2023{\natexlab{a}}.
\newblock \href {https://doi.org/10.18653/v1/2023.acl-long.910} {{RARR}: Researching and revising what language models say, using language models}.
\newblock In \emph{Proceedings of the 61st Annual Meeting of the Association for Computational Linguistics (Volume 1: Long Papers)}, pages 16477--16508, Toronto, Canada. Association for Computational Linguistics.

\bibitem[{Gao et~al.(2023{\natexlab{b}})Gao, Yen, Yu, and Chen}]{gao-etal-2023-enabling}
Tianyu Gao, Howard Yen, Jiatong Yu, and Danqi Chen. 2023{\natexlab{b}}.
\newblock \href {https://doi.org/10.18653/v1/2023.emnlp-main.398} {Enabling large language models to generate text with citations}.
\newblock In \emph{Proceedings of the 2023 Conference on Empirical Methods in Natural Language Processing}, pages 6465--6488, Singapore. Association for Computational Linguistics.

\bibitem[{Guu et~al.(2020)Guu, Lee, Tung, Pasupat, and Chang}]{pmlr-v119-guu20a}
Kelvin Guu, Kenton Lee, Zora Tung, Panupong Pasupat, and Ming{-}Wei Chang. 2020.
\newblock \href {http://proceedings.mlr.press/v119/guu20a.html} {Retrieval augmented language model pre-training}.
\newblock In \emph{Proceedings of the 37th International Conference on Machine Learning, {ICML} 2020, 13-18 July 2020, Virtual Event}, volume 119 of \emph{Proceedings of Machine Learning Research}, pages 3929--3938. {PMLR}.

\bibitem[{Honovich et~al.(2022)Honovich, Aharoni, Herzig, Taitelbaum, Kukliansy, Cohen, Scialom, Szpektor, Hassidim, and Matias}]{honovich-etal-2022-true-evaluating}
Or~Honovich, Roee Aharoni, Jonathan Herzig, Hagai Taitelbaum, Doron Kukliansy, Vered Cohen, Thomas Scialom, Idan Szpektor, Avinatan Hassidim, and Yossi Matias. 2022.
\newblock \href {https://doi.org/10.18653/v1/2022.naacl-main.287} {{TRUE}: Re-evaluating factual consistency evaluation}.
\newblock In \emph{Proceedings of the 2022 Conference of the North American Chapter of the Association for Computational Linguistics: Human Language Technologies}, pages 3905--3920, Seattle, United States. Association for Computational Linguistics.

\bibitem[{Jiang et~al.(2023{\natexlab{a}})Jiang, Sablayrolles, Mensch, Bamford, Chaplot, Casas, Bressand, Lengyel, Lample, Saulnier et~al.}]{jiang2023mistral}
Albert~Q Jiang, Alexandre Sablayrolles, Arthur Mensch, Chris Bamford, Devendra~Singh Chaplot, Diego de~las Casas, Florian Bressand, Gianna Lengyel, Guillaume Lample, Lucile Saulnier, et~al. 2023{\natexlab{a}}.
\newblock \href {https://arxiv.org/abs/2310.06825} {Mistral 7b}.
\newblock \emph{ArXiv preprint}, abs/2310.06825.

\bibitem[{Jiang et~al.(2024)Jiang, Qi, Hong, Fu, Cheng, Meng, Yu, Zhou, and Zhou}]{jiang2024large}
Che Jiang, Biqing Qi, Xiangyu Hong, Dayuan Fu, Yang Cheng, Fandong Meng, Mo~Yu, Bowen Zhou, and Jie Zhou. 2024.
\newblock \href {https://arxiv.org/abs/2403.20009} {On large language models' hallucination with regard to known facts}.
\newblock \emph{ArXiv preprint}, abs/2403.20009.

\bibitem[{Jiang et~al.(2023{\natexlab{b}})Jiang, Xu, Gao, Sun, Liu, Dwivedi-Yu, Yang, Callan, and Neubig}]{jiang-etal-2023-active}
Zhengbao Jiang, Frank Xu, Luyu Gao, Zhiqing Sun, Qian Liu, Jane Dwivedi-Yu, Yiming Yang, Jamie Callan, and Graham Neubig. 2023{\natexlab{b}}.
\newblock \href {https://doi.org/10.18653/v1/2023.emnlp-main.495} {Active retrieval augmented generation}.
\newblock In \emph{Proceedings of the 2023 Conference on Empirical Methods in Natural Language Processing}, pages 7969--7992, Singapore. Association for Computational Linguistics.

\bibitem[{Kadavath et~al.(2022)Kadavath, Conerly, Askell, Henighan, Drain, Perez, Schiefer, Hatfield-Dodds, DasSarma, Tran-Johnson et~al.}]{kadavath2022language}
Saurav Kadavath, Tom Conerly, Amanda Askell, Tom Henighan, Dawn Drain, Ethan Perez, Nicholas Schiefer, Zac Hatfield-Dodds, Nova DasSarma, Eli Tran-Johnson, et~al. 2022.
\newblock \href {https://arxiv.org/abs/2207.05221} {Language models (mostly) know what they know}.
\newblock \emph{ArXiv preprint}, abs/2207.05221.

\bibitem[{Kamath et~al.(2020)Kamath, Jia, and Liang}]{kamath-etal-2020-selective}
Amita Kamath, Robin Jia, and Percy Liang. 2020.
\newblock \href {https://doi.org/10.18653/v1/2020.acl-main.503} {Selective question answering under domain shift}.
\newblock In \emph{Proceedings of the 58th Annual Meeting of the Association for Computational Linguistics}, pages 5684--5696, Online. Association for Computational Linguistics.

\bibitem[{Lewis et~al.(2020)Lewis, Perez, Piktus, Petroni, Karpukhin, Goyal, K{\"{u}}ttler, Lewis, Yih, Rockt{\"{a}}schel, Riedel, and Kiela}]{lewis2020retrieval}
Patrick S.~H. Lewis, Ethan Perez, Aleksandra Piktus, Fabio Petroni, Vladimir Karpukhin, Naman Goyal, Heinrich K{\"{u}}ttler, Mike Lewis, Wen{-}tau Yih, Tim Rockt{\"{a}}schel, Sebastian Riedel, and Douwe Kiela. 2020.
\newblock \href {https://proceedings.neurips.cc/paper/2020/hash/6b493230205f780e1bc26945df7481e5-Abstract.html} {Retrieval-augmented generation for knowledge-intensive {NLP} tasks}.
\newblock In \emph{Advances in Neural Information Processing Systems 33: Annual Conference on Neural Information Processing Systems 2020, NeurIPS 2020, December 6-12, 2020, virtual}.

\bibitem[{Li et~al.(2023)Li, Patel, Vi{\'{e}}gas, Pfister, and Wattenberg}]{li2024inference}
Kenneth Li, Oam Patel, Fernanda~B. Vi{\'{e}}gas, Hanspeter Pfister, and Martin Wattenberg. 2023.
\newblock \href {http://papers.nips.cc/paper\_files/paper/2023/hash/81b8390039b7302c909cb769f8b6cd93-Abstract-Conference.html} {Inference-time intervention: Eliciting truthful answers from a language model}.

\bibitem[{Longpre et~al.(2021)Longpre, Perisetla, Chen, Ramesh, DuBois, and Singh}]{longpre-etal-2021-entity}
Shayne Longpre, Kartik Perisetla, Anthony Chen, Nikhil Ramesh, Chris DuBois, and Sameer Singh. 2021.
\newblock \href {https://doi.org/10.18653/v1/2021.emnlp-main.565} {Entity-based knowledge conflicts in question answering}.
\newblock In \emph{Proceedings of the 2021 Conference on Empirical Methods in Natural Language Processing}, pages 7052--7063, Online and Punta Cana, Dominican Republic. Association for Computational Linguistics.

\bibitem[{Mallen et~al.(2023)Mallen, Asai, Zhong, Das, Khashabi, and Hajishirzi}]{mallen-etal-2023-trust}
Alex Mallen, Akari Asai, Victor Zhong, Rajarshi Das, Daniel Khashabi, and Hannaneh Hajishirzi. 2023.
\newblock \href {https://doi.org/10.18653/v1/2023.acl-long.546} {When not to trust language models: Investigating effectiveness of parametric and non-parametric memories}.
\newblock In \emph{Proceedings of the 61st Annual Meeting of the Association for Computational Linguistics (Volume 1: Long Papers)}, pages 9802--9822, Toronto, Canada. Association for Computational Linguistics.

\bibitem[{Min et~al.(2023)Min, Krishna, Lyu, Lewis, Yih, Koh, Iyyer, Zettlemoyer, and Hajishirzi}]{min-etal-2023-factscore}
Sewon Min, Kalpesh Krishna, Xinxi Lyu, Mike Lewis, Wen-tau Yih, Pang Koh, Mohit Iyyer, Luke Zettlemoyer, and Hannaneh Hajishirzi. 2023.
\newblock \href {https://doi.org/10.18653/v1/2023.emnlp-main.741} {{FA}ct{S}core: Fine-grained atomic evaluation of factual precision in long form text generation}.
\newblock In \emph{Proceedings of the 2023 Conference on Empirical Methods in Natural Language Processing}, pages 12076--12100, Singapore. Association for Computational Linguistics.

\bibitem[{Niu et~al.(2024)Niu, Wu, Zhu, Xu, Shum, Zhong, Song, and Zhang}]{wu2023ragtruth}
Cheng Niu, Yuanhao Wu, Juno Zhu, Siliang Xu, KaShun Shum, Randy Zhong, Juntong Song, and Tong Zhang. 2024.
\newblock \href {https://doi.org/10.18653/v1/2024.acl-long.585} {{RAGT}ruth: A hallucination corpus for developing trustworthy retrieval-augmented language models}.
\newblock In \emph{Proceedings of the 62nd Annual Meeting of the Association for Computational Linguistics (Volume 1: Long Papers)}, pages 10862--10878, Bangkok, Thailand. Association for Computational Linguistics.

\bibitem[{Ram et~al.(2023)Ram, Levine, Dalmedigos, Muhlgay, Shashua, Leyton-Brown, and Shoham}]{ram-etal-2023-context}
Ori Ram, Yoav Levine, Itay Dalmedigos, Dor Muhlgay, Amnon Shashua, Kevin Leyton-Brown, and Yoav Shoham. 2023.
\newblock \href {https://doi.org/10.1162/tacl_a_00605} {In-context retrieval-augmented language models}.
\newblock \emph{Transactions of the Association for Computational Linguistics}, 11:1316--1331.

\bibitem[{Ren et~al.(2023)Ren, Luo, Zhao, Krishna, Saleh, Lakshminarayanan, and Liu}]{ren2022out}
Jie Ren, Jiaming Luo, Yao Zhao, Kundan Krishna, Mohammad Saleh, Balaji Lakshminarayanan, and Peter~J. Liu. 2023.
\newblock \href {https://openreview.net/pdf?id=kJUS5nD0vPB} {Out-of-distribution detection and selective generation for conditional language models}.
\newblock In \emph{The Eleventh International Conference on Learning Representations, {ICLR} 2023, Kigali, Rwanda, May 1-5, 2023}. OpenReview.net.

\bibitem[{Robertson and Walker(1994)}]{10.1007/978-1-4471-2099-5_24}
S.~E. Robertson and S.~Walker. 1994.
\newblock \href {https://dl.acm.org/doi/pdf/10.5555/188490.188561} {Some simple effective approximations to the 2-poisson model for probabilistic weighted retrieval}.
\newblock In \emph{SIGIR '94}, pages 232--241, London. Springer London.

\bibitem[{Schick et~al.(2023)Schick, Dwivedi{-}Yu, Dess{\`{\i}}, Raileanu, Lomeli, Hambro, Zettlemoyer, Cancedda, and Scialom}]{schick2024toolformer}
Timo Schick, Jane Dwivedi{-}Yu, Roberto Dess{\`{\i}}, Roberta Raileanu, Maria Lomeli, Eric Hambro, Luke Zettlemoyer, Nicola Cancedda, and Thomas Scialom. 2023.
\newblock \href {http://papers.nips.cc/paper\_files/paper/2023/hash/d842425e4bf79ba039352da0f658a906-Abstract-Conference.html} {Toolformer: Language models can teach themselves to use tools}.

\bibitem[{See et~al.(2017)See, Liu, and Manning}]{see-etal-2017-get}
Abigail See, Peter~J. Liu, and Christopher~D. Manning. 2017.
\newblock \href {https://doi.org/10.18653/v1/P17-1099} {Get to the point: Summarization with pointer-generator networks}.
\newblock In \emph{Proceedings of the 55th Annual Meeting of the Association for Computational Linguistics (Volume 1: Long Papers)}, pages 1073--1083, Vancouver, Canada. Association for Computational Linguistics.

\bibitem[{Shao et~al.(2023)Shao, Gong, Shen, Huang, Duan, and Chen}]{shao-etal-2023-enhancing}
Zhihong Shao, Yeyun Gong, Yelong Shen, Minlie Huang, Nan Duan, and Weizhu Chen. 2023.
\newblock \href {https://doi.org/10.18653/v1/2023.findings-emnlp.620} {Enhancing retrieval-augmented large language models with iterative retrieval-generation synergy}.
\newblock In \emph{Findings of the Association for Computational Linguistics: EMNLP 2023}, pages 9248--9274, Singapore. Association for Computational Linguistics.

\bibitem[{Shi et~al.(2024{\natexlab{a}})Shi, Han, Lewis, Tsvetkov, Zettlemoyer, and Yih}]{shi2023trusting}
Weijia Shi, Xiaochuang Han, Mike Lewis, Yulia Tsvetkov, Luke Zettlemoyer, and Wen-tau Yih. 2024{\natexlab{a}}.
\newblock \href {https://doi.org/10.18653/v1/2024.naacl-short.69} {Trusting your evidence: Hallucinate less with context-aware decoding}.
\newblock In \emph{Proceedings of the 2024 Conference of the North American Chapter of the Association for Computational Linguistics: Human Language Technologies (Volume 2: Short Papers)}, pages 783--791, Mexico City, Mexico. Association for Computational Linguistics.

\bibitem[{Shi et~al.(2024{\natexlab{b}})Shi, Min, Lomeli, Zhou, Li, Lin, Smith, Zettlemoyer, Yih, and Lewis}]{shi2023context}
Weijia Shi, Sewon Min, Maria Lomeli, Chunting Zhou, Margaret Li, Xi~Victoria Lin, Noah~A. Smith, Luke Zettlemoyer, Wen{-}tau Yih, and Mike Lewis. 2024{\natexlab{b}}.
\newblock \href {https://openreview.net/forum?id=LXVswInHOo} {In-context pretraining: Language modeling beyond document boundaries}.
\newblock In \emph{The Twelfth International Conference on Learning Representations, {ICLR} 2024, Vienna, Austria, May 7-11, 2024}. OpenReview.net.

\bibitem[{Shi et~al.(2024{\natexlab{c}})Shi, Min, Yasunaga, Seo, James, Lewis, Zettlemoyer, and Yih}]{shi2023replug}
Weijia Shi, Sewon Min, Michihiro Yasunaga, Minjoon Seo, Richard James, Mike Lewis, Luke Zettlemoyer, and Wen-tau Yih. 2024{\natexlab{c}}.
\newblock \href {https://doi.org/10.18653/v1/2024.naacl-long.463} {{REPLUG}: Retrieval-augmented black-box language models}.
\newblock In \emph{Proceedings of the 2024 Conference of the North American Chapter of the Association for Computational Linguistics: Human Language Technologies (Volume 1: Long Papers)}, pages 8371--8384, Mexico City, Mexico. Association for Computational Linguistics.

\bibitem[{Shrivastava et~al.(2023)Shrivastava, Liang, and Kumar}]{shrivastava2023llamas}
Vaishnavi Shrivastava, Percy Liang, and Ananya Kumar. 2023.
\newblock \href {https://arxiv.org/abs/2311.08877} {Llamas know what gpts don't show: Surrogate models for confidence estimation}.
\newblock \emph{ArXiv preprint}, abs/2311.08877.

\bibitem[{Tang et~al.(2024)Tang, Laban, and Durrett}]{tang2024minicheck}
Liyan Tang, Philippe Laban, and Greg Durrett. 2024.
\newblock Minicheck: Efficient fact-checking of llms on grounding documents.
\newblock \emph{arXiv preprint arXiv:2404.10774}.

\bibitem[{Touvron et~al.(2023)Touvron, Martin, Stone, Albert, Almahairi, Babaei, Bashlykov, Batra, Bhargava, Bhosale et~al.}]{touvron2023llama}
Hugo Touvron, Louis Martin, Kevin Stone, Peter Albert, Amjad Almahairi, Yasmine Babaei, Nikolay Bashlykov, Soumya Batra, Prajjwal Bhargava, Shruti Bhosale, et~al. 2023.
\newblock \href {https://arxiv.org/abs/2307.09288} {Llama 2: Open foundation and fine-tuned chat models}.
\newblock \emph{ArXiv preprint}, abs/2307.09288.

\bibitem[{Trivedi et~al.(2023)Trivedi, Balasubramanian, Khot, and Sabharwal}]{trivedi-etal-2023-interleaving}
Harsh Trivedi, Niranjan Balasubramanian, Tushar Khot, and Ashish Sabharwal. 2023.
\newblock \href {https://doi.org/10.18653/v1/2023.acl-long.557} {Interleaving retrieval with chain-of-thought reasoning for knowledge-intensive multi-step questions}.
\newblock In \emph{Proceedings of the 61st Annual Meeting of the Association for Computational Linguistics (Volume 1: Long Papers)}, pages 10014--10037, Toronto, Canada. Association for Computational Linguistics.

\bibitem[{Wang et~al.(2024)Wang, Song, Peng, Tian, Jin, Mi, Su, and Yu}]{wang2024fine}
Ante Wang, Linfeng Song, Baolin Peng, Ye~Tian, Lifeng Jin, Haitao Mi, Jinsong Su, and Dong Yu. 2024.
\newblock \href {https://arxiv.org/abs/2402.15631} {Fine-grained self-endorsement improves factuality and reasoning}.
\newblock \emph{ArXiv preprint}, abs/2402.15631.

\bibitem[{Wu et~al.(2024)Wu, Wu, Cassasola, Zhang, Wei, Nguyen, Riantawan, Riantawan, Ho, and Zou}]{wu2024well}
Kevin Wu, Eric Wu, Ally Cassasola, Angela Zhang, Kevin Wei, Teresa Nguyen, Sith Riantawan, Patricia~Shi Riantawan, Daniel~E Ho, and James Zou. 2024.
\newblock \href {https://arxiv.org/abs/2402.02008} {How well do llms cite relevant medical references? an evaluation framework and analyses}.
\newblock \emph{ArXiv preprint}, abs/2402.02008.

\bibitem[{Xie et~al.(2023)Xie, Zhang, Chen, Lou, and Su}]{xie2023adaptive}
Jian Xie, Kai Zhang, Jiangjie Chen, Renze Lou, and Yu~Su. 2023.
\newblock \href {https://openreview.net/forum?id=auKAUJZMO6} {Adaptive chameleon or stubborn sloth: Revealing the behavior of large language models in knowledge conflicts}.
\newblock In \emph{The Twelfth International Conference on Learning Representations}.

\bibitem[{Xu et~al.(2024)Xu, Shi, and Choi}]{xu2023recomp}
Fangyuan Xu, Weijia Shi, and Eunsol Choi. 2024.
\newblock \href {https://openreview.net/forum?id=mlJLVigNHp} {{RECOMP:} improving retrieval-augmented lms with context compression and selective augmentation}.
\newblock In \emph{The Twelfth International Conference on Learning Representations, {ICLR} 2024, Vienna, Austria, May 7-11, 2024}. OpenReview.net.

\bibitem[{Yelp(2021)}]{yelpopendataset}
Yelp. 2021.
\newblock \href {https://www.yelp.com/dataset} {Yelp open dataset}.

\bibitem[{Yin et~al.(2024)Yin, Srinivasa, and Chang}]{yin2024characterizing}
Fan Yin, Jayanth Srinivasa, and Kai{-}Wei Chang. 2024.
\newblock \href {https://openreview.net/forum?id=7DbIyQlfaO} {Characterizing truthfulness in large language model generations with local intrinsic dimension}.
\newblock In \emph{Forty-first International Conference on Machine Learning, {ICML} 2024, Vienna, Austria, July 21-27, 2024}. OpenReview.net.

\bibitem[{Zha et~al.(2023)Zha, Yang, Li, and Hu}]{zha-etal-2023-alignscore}
Yuheng Zha, Yichi Yang, Ruichen Li, and Zhiting Hu. 2023.
\newblock \href {https://doi.org/10.18653/v1/2023.acl-long.634} {{A}lign{S}core: Evaluating factual consistency with a unified alignment function}.
\newblock In \emph{Proceedings of the 61st Annual Meeting of the Association for Computational Linguistics (Volume 1: Long Papers)}, pages 11328--11348, Toronto, Canada. Association for Computational Linguistics.

\bibitem[{Zhang et~al.(2023)Zhang, Li, Cui, Cai, Liu, Fu, Huang, Zhao, Zhang, Chen et~al.}]{zhang2023siren}
Yue Zhang, Yafu Li, Leyang Cui, Deng Cai, Lemao Liu, Tingchen Fu, Xinting Huang, Enbo Zhao, Yu~Zhang, Yulong Chen, et~al. 2023.
\newblock \href {https://arxiv.org/abs/2309.01219} {Siren's song in the ai ocean: a survey on hallucination in large language models}.
\newblock \emph{ArXiv preprint}, abs/2309.01219.

\bibitem[{Zhou et~al.(2023)Zhou, Zhang, Poon, and Chen}]{zhou-etal-2023-context}
Wenxuan Zhou, Sheng Zhang, Hoifung Poon, and Muhao Chen. 2023.
\newblock \href {https://doi.org/10.18653/v1/2023.findings-emnlp.968} {Context-faithful prompting for large language models}.
\newblock In \emph{Findings of the Association for Computational Linguistics: EMNLP 2023}, pages 14544--14556, Singapore. Association for Computational Linguistics.

\end{thebibliography}

\clearpage
\appendix
\twocolumn[{%
 \centering
 \Large\bf Supplementary Material: Appendices \\ [20pt]
}]
\label{sec:appendix}

\section{\modelname Details}
\label{appendix-feature-impl}

\subsection{Decoding-time feature collection}

\setlength{\tabcolsep}{3.5pt}
\begin{table}[ht!]
  \centering
  \resizebox{\linewidth}{!}{
  \begin{tabular}{c | c | c }
    \hline
    \textbf{Feature Type} & \textbf{Alias} & \textbf{Range} \\
    \hline
    \multirow{2}{*}{\textbf{Likelihood}} & \texttt{min\_prob} & $[0, 1]$\\
     & \texttt{mean\_prob} & $[0, 1]$\\
    \hdashline
    \multirow{5}{*}{\textbf{Uncertainty}} & \texttt{max\_entropy} & $[0, 1]$\\
     & \texttt{mean\_entropy} & $[0, 1]$\\
     & \texttt{lid\_layer\_15} & $[0, +\infty)$\\
     & \texttt{lid\_layer\_16} & $[0, +\infty)$\\
     & \texttt{lid\_layer\_17} & $[0, +\infty)$\\
     \hdashline
    \multirow{2}{*}{\textbf{Context Influence}} & \texttt{mean\_contrastive\_kl} & $[0, 1]$\\
     & \texttt{large\_kl\_pos} & $[0, +\infty)$\\
    \hdashline
    \textbf{Semantic Alignment} & \texttt{align\_score} & $[0, 1]$\\
  \hline
  \end{tabular}
  }
  \caption{A list of features monitored by \modelname. }
  \label{tab:feature_details}
  
\end{table}

In Table \ref{tab:feature_details}, we outline the full list of features monitored and leveraged by \modelname. In the rest of this section, we detail the methodology used to measure each feature.

\paragraph{\texttt{min\_prob} and \texttt{mean\_prob}} For each of the predicted token in $s_i$, we calculate its likelihood. \texttt{min\_prob} takes the likelihood of the most unlikely token and \texttt{mean\_prob} takes the average.

\paragraph{\texttt{mean\_entropy} and \texttt{max\_entropy}} For each sentence $s_i$, we calculate the entropy of the distribution over the entire vocabulary for each of the output token position. \texttt{mean\_entropy} takes the average of these values and \texttt{max\_entropy} takes the max.

\paragraph{\texttt{lid\_layer\_x}} When the model finishes decoding for a single segment $S=[s_0, s_1, ..., s_n]$, we collect the hidden representation of the last token $s_n$ from layer \texttt{x}. Then, we calculate the local intrinsic dimension using the representation against a set of T pre-computed hidden dimensions measured in the same way using labelled segments from the train set, following the LID-MLE method proposed in \citet{yin2024characterizing}. Specifically, we sort the set of T reference points by their distance to $s_n$ and calculate local intrinsic dimension as 
$LID(S) = \frac{1}{T-1}(\sum_{j=1}^T \log (D(s_n, q_T) / D(s_n, q_j)))^{-1}$, where $D(s_n, q_j)$ is the Euclidean distance from the $s_n$ representation to the $j$-th reference point. Untruthful sentences will have larger LID values. In preliminary studies, we found  \texttt{lid\_layer\_x} to be ineffective for Summ and Data2txt. Therefore, we only incorporate this set of features for FS, F-100, F-100-anti, and QA.

\paragraph{\texttt{mean\_contrastive\_kl} and \texttt{large\_kl\_pos}} As the model decodes each token in the sentence $S=[s_0, s_1, ..., s_n]$, we simultaneously collect the distribution over the vocabulary as well as the distribution with the same generation prefix but without the retrieved context. Concretely, given the LLM $\mathcal{M}$, a query $X$, a context $C$, a output prefix $O$, then the two distributions could be written as $D_{i1} = P_\mathcal{M}(s_i|X; C; O; [s_0, …, s_{i-1}])$ and $D_{i2} = P_\mathcal{M}(s_i|X; O; [s_0, …, s_{i-1}])$. Then, we obtain a sequence of KL-divergence values $KL(D_{i1}||D_{i2})$ for S. The feature \texttt{mean\_contrastive\_kl} the average across all positions in this sentence. The feature \texttt{mean\_contrastive\_kl} the average across all positions in this sentence. \texttt{large\_kl\_pos} is a discrete version of the mean signal which counts the number of positions with the KL-divergence greater than 3.0. 

\paragraph{\texttt{align\_score}} For each sentence, we calculate its AlignScore \citep{zha-etal-2023-alignscore} with the context as the reference. We use the AlignScore-base model released by the authors for its efficiency.

For all the feature collection, classification, and decoding experiments, we execute them on a local server with 8 Nvidia A6000 GPUs. For models of size 7B, the latency of running the full decoding pipeline is roughly 30s per sample using two GPUs. To run the end-to-end decoding and evaluation on the test set, the latency is roughly 2 hours per task. 

\subsection{Aggregator training details}

We use the \texttt{scikit-learn} and the \texttt{xgboost} Python package to train different variations of \modelname. For \textsc{SynCheck}\textsubscript{MLP}, we use a two-layer neural network with hidden dimension 100 and ReLU activation. We use a learning rate of 0.001, Adam optimizer, batch size 128, and train for 300 iterations. While we did not extensively tune the hyperparameters, we did a preliminary study with learning rate \{0.01, 0.001, 0.0001\} and batch size \{32, 64, 128\} which supports the chosen set of hyperparameters.

\section{Benchmark Construction Details}
\label{appendix-benchmark-details}

\subsection{F-100 and F-100-anti}

To create F-100 and F-100-anti, we first directly prompt \texttt{ChatGPT} to propose 100 entities representing famous people. Figure \ref{famous-entities} presents a full list of the entities. Then, we run BM25 retrieval \citep{10.1007/978-1-4471-2099-5_24} on paragraph-chunked Wikipedia data (version 2021-10-13, processed by \citet{chen2023dense}) with \texttt{"Tell me a bio of [entity]"} as the query. For F-100, we keep and use the top-10 paragraphs as the retrieved evidence. For F-100-anti, we link each entity E1 with the contexts retrieved for another entity E2, while replacing all the occurrences of E2 with E1. 

\setlength{\tabcolsep}{3.5pt}
\begin{table*}[ht!]
  \centering
  \resizebox{\linewidth}{!}{
  \begin{tabular}{c | c | c | c | c | c | c | c }
    \hline
    \textbf{Dataset} & \textbf{Split} & \textbf{\# Docs} & \textbf{\# Instances} & \textbf{\# Cxt Sents} & \textbf{\# Gen Sents} & \textbf{\% Faithful Sents} & \textbf{\% Faithful Instances} \\
    \hline
    \multicolumn{8}{c}{\textbf{\texttt{Llama 2 7B Chat}}} \\
    \hline
    \multirow{2}{*}{\textbf{QA}} & train & 839 & 9669 & 17.7 & 14.98 & 87.06\% & 54.47\% \\
    & test & 150 & 1453 & 17.1 & 13.34 & 90.71\% & 34.67\% \\
    \hdashline
    \multirow{2}{*}{\textbf{Summ}} & train & 793 & 3771 & 35.9 & 5.43 & 88.17\% & 48.30\% \\
    & test & 150 & 692 & 34.8 & 5.05 & 91.47\% & 33.33\% \\
    \hdashline
    \multirow{2}{*}{\textbf{Data2txt}} & train & 883 & 5750 & 23.8 & 7.13 & 77.51\% & 86.64\% \\
    & test & 150 & 940 & 24.3 & 6.65 & 80.53\% & 82.00\%  \\
    \hdashline
    \textbf{FS} & test & 500 & 1770 & 51.5 & 4.46 & 73.62\% & 51.40\%     \\
    \textbf{F-100} & test & 100 & 405 & 38.2 & 4.28 & 56.79\% & 67.00\%  \\
    \textbf{F-100-anti} & test & 100 & 349 & 36.6 & 4.01 & 53.58\% & 73.00\% \\
    \hline
    \multicolumn{8}{c}{\textbf{\texttt{Mistral 7B Instruct}}} \\
    \hline
    \multirow{2}{*}{\textbf{QA}} & train & 839 & 6055 & 18.6 & 11.99 & 88.16\% & 41.24\% \\
    & test & 150 & 882 & 18.1 & 10.60 & 91.84\% & 20.67\%  \\
    \hdashline
    \multirow{2}{*}{\textbf{Summ}} & train & 793 & 4530 & 33.9 & 6.40 & 84.55\% & 66.83\%  \\
    & test & 150 & 838 & 33.0 & 6.13 & 86.52\% & 57.33\%   \\
    \hdashline
    \multirow{2}{*}{\textbf{Data2txt}} & train & 883 & 7677 & 23.0 & 9.38 & 79.77\% & 93.32\%  \\
    & test & 150 & 1269 & 23.5 & 9.13 & 82.74\% & 88.67\%  \\
    \hdashline
    \textbf{FS} & test & 290 & 1629 & 51.9 & 6.37 & 83.00\% & 58.28\%  \\
    \textbf{F-100} & test & 61 & 227 & 38.0 & 3.94 & 79.30\% & 40.98\%  \\
    \textbf{F-100-anti} & test & 62 & 214 & 34.8 & 3.76 & 55.14\% & 62.90\% \\
  \hline
  \end{tabular}
  }
  \caption{Basic statistics of our evaluation benchmark. Notably, our benchmark tests long-form generation with long-form retrieved evidence. \# Cxt Sents and \# Gen Sents stand for the number of sentences in the context and the model's output. Both the tested models exhibit a considerable rate of unfaithful sentences.}
  \label{tab:dataset_details}
  
\end{table*}

\begin{figure}[h!]
\small
\centering
\begin{tabular}{p{0.98\linewidth}}
    \hline
    \texttt{Nikita Khrushchev, Malcolm X, John F. Kennedy, Bill Gates, Queen Elizabeth II, Napoleon Bonaparte, Mohandas Gandhi, Albert Einstein, Cleopatra, Adolf Hitler, Freddie Mercury, Plato, J.K. Rowling, Karl Marx, Margaret Thatcher, Angela Merkel, Vincent Van Gogh, John Lennon, Che Guevara, Cristiano Ronaldo, Rosa Parks, Rihanna, Alexander the Great, Isaac Newton, Julius Caesar, Amelia Earhart, Simone Biles, Michael Jordan, Elton John, Sigmund Freud, Joseph Stalin, Jane Goodall, Beyoncé, Adele, Charles Dickens, Thomas Edison, Eminem, Virginia Woolf, Taylor Swift, Tupac Shakur, Justin Bieber, Tim Berners-Lee, Kanye West, Marie Curie, Ludwig van Beethoven, Pablo Neruda, Steve Jobs, Usain Bolt, Bruce Lee, Marilyn Monroe, Michael Jackson, Shakira, Mozart, Lady Gaga, Vladimir Putin, Charles Darwin, Harriet Tubman, Benjamin Franklin, Oprah Winfrey, Malala Yousafzai, Socrates, Mahatma Gandhi, Pablo Picasso, Frida Kahlo, Nelson Mandela, Whitney Houston, Winston Churchill, Stephen Hawking, Fidel Castro, Margaret Atwood, Madonna, Leonardo DiCaprio, Elizabeth I, Galileo Galilei, Muhammad Ali, Mao Zedong, William Shakespeare, Joan of Arc, George Washington, Mikhail Gorbachev, Abraham Lincoln, Pele, Martin Luther King Jr., Jane Austen, Ed Sheeran, Sachin Tendulkar, Ariana Grande, Wolfgang Amadeus Mozart, Mark Zuckerberg, Vincent van Gogh, Bob Marley, Ronald Reagan, Barack Obama, Britney Spears, Walt Disney, Leonardo da Vinci, Elvis Presley, Lionel Messi, Anne Frank, Confucius} \\
    \hline
\end{tabular}
\caption{A list of the well-known entities included in the F-100 and F-100-anti biography generation dataset.}
\label{famous-entities}
\end{figure}

\subsection{Trustworthiness label generation}
\label{appendix-trustworthy-label-generation}

In this section, we describe the lexical matching procedure we design for obtaining sentence-level faithfulness label. 

\paragraph{RAGTruth} For QA, Summ, and Data2txt, the hallucination spans annotated by RAGTruth are in the format (start, end) indication the character-level boundary. For each sentence in the output, the sentence is then labelled as unfaithful if either (1) it contains one or more hallucination spans or (2) it is contained by a hallucination span. 

\paragraph{Biography} For FS, F-100, and F-100-anti, the outputs are converted to propositions and judged by the AutoAIS system\footnote{We use the model released at \url{https://huggingface.co/google/t5_xxl_true_nli_mixture}.}. We map the propositions back to sentence-level labels using token recall. Specifically, for each unfaithful proposition, we find the sentence that contains the highest proportion of its tokens and label it as unfaithful. The rest sentences are labelled as faithful. We conduct a small-scale human inspection and find that this approach generally produces correct label mappings.

\subsection{Dataset Statistics}

We further outline the dataset statistics in Table \ref{tab:dataset_details}, including the instance count, retrieval length, generation length, as well as the label distribution.

\section{Implementation Details of Baselines}
\label{appendix-baseline-implementation}

In this section, we further describe the implementation details of two context faithfulness tarcking baselines we compare with \modelname.

\paragraph{\textsc{SpanExtract}} For QA, Summ, and Data2txt, we follow the evaluation prompt in the original paper \citep{wu2023ragtruth}. For the biography generation tasks, we use the following evaluation prompt, which share some similarity with the original prompt for evaluating QA.

\begin{center}
   \texttt{Passages: \{Retrieved Context\}\\Question: Tell me a bio of \{entity\}.\\Answer: \{answer\}\\Please analyze the provided Passages, Question, and Answer, and indicate whether there are any hallucinated (invented or incorrect) parts in the answer. If there are, please specify the exact span that is hallucinated. Please provide a response in the following format:...(json format) ...}
\end{center}

\paragraph{\textsc{CriticTok}} For Self-RAG, we refer to the original implementation and use the following prefix for scoring each segment:

\begin{center}
    \texttt{\{Instruction\} \textcolor{blue}{[Retrieve]} \{Retrieved Context\} \textcolor{blue}{[Relevant]} \{Response Segment\}}
\end{center}

We also tried using the full response until the segment instead of only a single segment. However, we find that the previous approach performs slightly better and thus select it as the baseline.

\begin{center}
    \texttt{\{Instruction\} \textcolor{blue}{[Retrieve]} \{Retrieved Context\} \textcolor{blue}{[Relevant]} \{Full Response until Segment\}}
\end{center}

\section{Further Analysis}
\subsection{Single-feature reranking baselines}
\label{appendix-more-rerank-baselines}

To further verify the effectiveness of the proposed detection and decoding algorithms, we extend \cref{tab:intervention_main_results} with two additional baselines: reranking with either only semantic alignment or only likelihood. These baselines are directly comparable with the line “Reranking\ding{95}” in \cref{tab:intervention_main_results}, which is copied to the third line of \cref{tab:intervention_further_results}. As we can observe, reranking with \modelname outperforms reranking with these individual features. The full \searchalgname algorithm further improves the output’s faithfulness.

\setlength{\tabcolsep}{3.5pt}
\begin{table*}[ht!]
 \centering
 \resizebox{0.9\linewidth}{!}{
 \begin{tabular}{c | c c | c c | c c | c c | c c }
 \hline
 \multirow{2}{*}{\textbf{Method}} & \multicolumn{2}{c|}{\textbf{QA}} & \multicolumn{2}{c|}{\textbf{Summ}} & \multicolumn{2}{c|}{\textbf{Data2txt}} & \multicolumn{2}{c|}{\textbf{F-100}}  & \multicolumn{2}{c}{\textbf{Average}}\\
 & Faith. & Info. & Faith. & Info. & Faith. & Info. & Faith. & Info. & Faith. & Info.  \\
 \hline
 \multicolumn{11}{c}{\textbf{\texttt{Llama 2 7B Chat}}} \\
 \hdashline
 Reranking (Likelihood) & 0.686 & 8.5 & 0.731 & 9.5 & 0.468 & 14.1 & 0.513 & 11.9 & 0.600 & 11.0  \\
 Reranking (AlignScore) &  0.738 & 8.8 & 0.731 & 9.3 & 0.468 & 13.3 & 0.538 & 9.7 & 0.619 & 10.3 \\
 Reranking\textsuperscript{\ding{95}} & 0.745 & 9.3 & 0.757 & 9.8 & 0.474 & 13.4 & 0.563 & 10.0 & 0.635 & 0.106  \\
 \searchalgname (Full)\textsuperscript{\ding{95}} & \textbf{0.768} & 4.3 & \textbf{0.770} & 9.5 & \textbf{0.529} & 10.4 & \textbf{0.735} & 0.48 & \textbf{0.701} & 7.3 \\
 \hline
 \end{tabular}
 }
 \caption{Further Faithfulness-Informativeness evaluation results of faithfulness intervention methods. We further compare the best results in \cref{tab:intervention_main_results} with two baselines of reranking with a single feature. We use \ding{95} to mark the decoding methods that leverage \modelname.} 
 \label{tab:intervention_further_results}
\end{table*}

\subsection{Hyperparameter for \searchalgname}
\label{appendix-search-hyperparam}

In this section, we provide further studies on the hyperparameter sensitivity for \searchalgname. Specifically, we use the Llama 2 7B Chat model and investigate two hyperparameters: the beam size $B$ and the sample size $S$. We keep the other parameters the same as unsed in the main experiments.

\begin{figure*}[t!]
\centering
\includegraphics[width=0.97\textwidth]{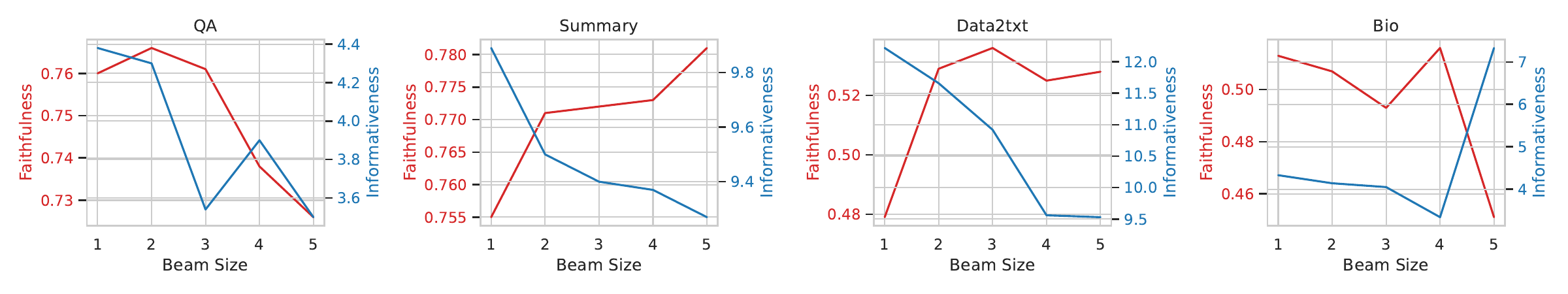}
\caption{Faithfulness and Informativeness of Llama 2 7B Chat as a function of the beam size.}
\label{fod=hyperparam-beam-size}
\end{figure*}

\begin{figure*}[t!]
\centering
\includegraphics[width=0.97\textwidth]{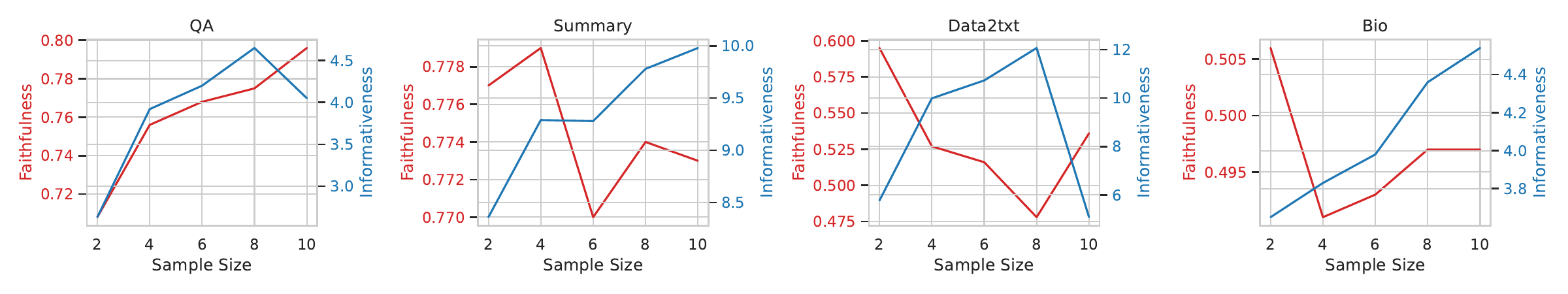}
\caption{Faithfulness and Informativeness of Llama 2 7B Chat as a function of the sample size.}
\label{fod=hyperparam-sample-size}
\end{figure*}

We present the results in Figure \ref{fod=hyperparam-beam-size} and Figure \ref{fod=hyperparam-sample-size}. For the beam size, we observe that a higher beam size generally decreases the informativeness. This could be due to more easily achieving the end-of-sequence token with more diverse exploration patterns. Meanwhile, the faithfulness change under beam size changes varies task-by-task. For the sample size, we find a positive correlation with the informativeness and an U-shaped behavior for faithfulness. Generally, combination of modest sample size and modest beam size works well.

\subsection{Qualitative Study}
\label{appendix-qualitative-study}

In Figure \ref{qualitative-study-selection}, we present examples collected from QA, Summ, Data2txt, and FS using Llama 2 7B Chat. SynCheck provides an accurate detection of all the unfaithful sentences. In addition, it is able to recognize the faithful sentences following unfaithful ones, enabling the lookahead and the selection interventions after backtracking. 

\begin{figure*}[h!]
\small
\centering
\begin{tabular}{p{0.98\linewidth}}
    \hline
    \textbf{Instruction:} Republicans use \_\_\_\_\_\_\_\_ to allocate delegates in their presidential nominating procedure quizlet \\
    \textbf{Context:} [1] Some states award delegates to candidates on a winner-take-all basis, meaning that the candidate with the most votes in a state is awarded all of that state's delegates. Other states award delegates in proportion to each candidate's share of the primary vote. [2] To win the nomination, a candidate must win the votes of at least 1,191 delegates at the convention. Sen. John McCain (R-AZ) reached that benchmark in March 2008, becoming the Republican Party's presumptive nominee. [3] Most will have already pledged to whichever candidate won their state. To win the nomination, a candidate must win the votes of at least 1,191 delegates at the convention. Sen. John McCain (R-AZ) reached that benchmark in March 2008, becoming the Republican Party's presumptive nominee. \\
    \textbf{Response:} \textcolor{blue}{Based on the provided passages, Republicans use a proportional method to allocate delegates in their presidential nominating procedure.} [\texttt{SynCheck: 0.849}] \textcolor{blue}{According to passage 1, some states award delegates to candidates on a winner-take-all basis, while others award delegates in proportion to each candidate's share of the primary vote.} [\texttt{SynCheck: 0.964}] \textcolor{red}{This suggests that the allocation of delegates is not solely determined by a simple majority vote, but rather takes into account the distribution of support among the various candidates.} [\texttt{SynCheck: 0.302}] \textcolor{red}{Therefore, the answer to the question is: Proportional.} [\texttt{SynCheck: 0.827}] \\
    \hline     
    \textbf{Instruction:} Summarize the following news within 53 words \\
    \textbf{Context:} Their relationship led to jail time for her, but Mary Kay Letourneau Fualaau wants the world to know that she and her husband, Vili Fualaau, are still happily together. She was a married 34-year-old teacher and mother of four in Seattle in 1996 when she began an affair with Fualaau, her 13-year-old student. Letourneau gave birth to her young lover's child and went on to serve more than seven years in prison on charges related to their sexual relationship. On Friday, ABC's "20/20" will air an exclusive interview with the couple, who will celebrate their 10th wedding anniversary on May 20. The pair wed soon after she was released from prison in 2005 and are now the parents of two teen girls. According to ABC, during the interview with Barbara Walters, "Mary Kay tells Walters what makes their marriage work in spite of their huge difference in their age." "She also discusses her surprising plans to teach again, as well as her intentions to have her status as a registered sex offender lifted," a posting on the network's site says. "Vili Fualaau, meanwhile, discusses his bouts with alcoholism, depression and why he believes the system failed him while he was still a minor." Letourneau Fualaau is now 53, and her husband is 31.\\
    \textbf{Response:} \textcolor{blue}{Mary Kay Letourneau, who served 7 years in prison for having an affair with her 13-year-old student, is still married to her husband Vili Fualaau and they are celebrating their 10th wedding anniversary this May.} [\texttt{SynCheck: 0.974}] \textcolor{red}{Despite their large age gap, they claim their marriage works due to their shared values and love for each other.} [\texttt{SynCheck: 0.627}] \textcolor{blue}{Letourneau hopes to become a teacher again and has applied to have her sex offender status lifted.} [\texttt{SynCheck: 0.945}] \\
    \hline 
    \textbf{Instruction:} Write an objective overview about the following local business based only on the provided structured data. \\
    \textbf{Context:} \{'name': 'Super Cucas', 'address': '6527 Madrid Rd', 'city': 'Isla Vista', 'state': 'CA', 'categories': 'Ethnic Food, Specialty Food, Local Flavor, Mexican, Restaurants, Food, Breakfast \& Brunch, Fast Food, Juice Bars \& Smoothies', 'hours': \{'Monday': '7:0-0:0', 'Tuesday': '7:0-0:0', 'Wednesday': '7:0-0:0', 'Thursday': '7:0-0:0', 'Friday': '7:0-0:0', 'Saturday': '7:0-0:0', 'Sunday': '7:0-0:0'\}, 'attributes': \{'BusinessParking': \{'garage': False, 'street': True, 'validated': False, 'lot': None, 'valet': False\}, 'RestaurantsReservations': False, 'OutdoorSeating': None, 'WiFi': 'no', 'RestaurantsTakeOut': True, 'RestaurantsGoodForGroups': True, 'Music': None, 'Ambience': \{'touristy': False, 'hipster': False, 'romantic': False, 'divey': None, 'intimate': False, 'trendy': False, 'upscale': False, 'classy': False, 'casual': True\}\}, 'business\_stars': 3.5, 'review\_info': [\{'review\_stars': 4.0, 'review\_date': '2021-11-03 21:59:35', 'review\_text': "Super Cucas is really tasty and the portion sizes are so generous. They're also open late which is always great when you're in a college town. I love that they have a menu that offers every style of meat veganized (including soy pastor!) I'd love to try the vegan fries with meat (asada fries) and the breakfast burritos on my next visit. The portions are huge so I would recommend splitting or saving for 2 meals."\}]\}\\
    \textbf{Response:} \textcolor{blue}{Super Cucas is a locally-owned restaurant located in Isla Vista, California, serving a variety of Ethnic, Specialty, and Mexican cuisine.} [\texttt{SynCheck: 0.939}] \textcolor{red}{With a rating of 3.5 stars, the restaurant is popular among locals and visitors alike, particularly for its late hours of operation (until 10 PM from Monday to Saturday).} [\texttt{SynCheck: 0.594}] \textcolor{blue}{The establishment is known for its generous portion sizes and vegetarian options, including veganized meats.} [\texttt{SynCheck: 0.988}] \textcolor{blue}{While some reviewers have praised the quality of the food, others have experienced poor customer service, with instances of incorrect orders and rude staff behavior.} [\texttt{SynCheck: 0.801}] \textcolor{red}{Nonetheless, Super Cucas remains a popular spot for those seeking tasty and affordable meals in the area.} [\texttt{SynCheck: 0.794}]
    \\
    \hline
    \textbf{Instruction:} Who is Blanca Errázuriz? \\
    \textbf{Context:} [1] Blanca Errázuriz Blanca Errázuriz Blanca Elena Errázuriz Vergara (9 April 1894 - 20 March 1940), also known as Bianca de Saulles, was a Chilean socialite and the former wife of football player and businessman John de Saulles. In August 1917, Errázuriz fatally shot de Saulles multiple times during a disagreement over the custody of their son. After a highly publicized and sensational trial, Errázuriz was acquitted of her ex-husband's murder. Errázuriz was born in Via del Mar, Chile, the eldest daughter of Guillermo Errázuriz Urmeneta and of Blanca Vergara Alvarez, a beauty known as the Star of Santiago. [2] Blanca Errázuriz press, and the champion of the suffragettes who portrayed her as the victim of the chauvinism prevalent in the society of the time. Blanca Errázuriz was unanimously acquitted of the murder charges on 1 December 1917 in what was called a "popular" verdict. Following the trial, Errázuriz moved to San Francisco where she sought and obtained full custody of her son. She and her son went later to Japan. Eventually she and her son returned to Chile and settled there. On 22 December 1921, she remarried, this time to engineer Fernando Santa Cruz Wilson in Santiago; the couple later divorced. ... (8 documents omitted) \\
    \textbf{Response:} \textcolor{blue}{Blanca Errázuriz was a Chilean socialite and the former wife of football player and businessman John de Saulles.} [\texttt{SynCheck: 0.954}] \textcolor{red}{She was born in Viña del Mar, Chile in 1894 and was known for her beauty and her marriage to de Saulles, who was 15 years her senior.} [\texttt{SynCheck: 0.033}] \textcolor{blue}{In 1917, Errázuriz fatally shot de Saulles during a disagreement over the custody of their son, and she was acquitted of his murder after a highly publicized and sensational trial.} [\texttt{SynCheck: 0.912}] \textcolor{red}{Errázuriz was also known for her friendship with Rudolph Valentino and her association with the politically influential Errázuriz family, of Basque descent.} [\texttt{SynCheck: 0.654}]
  \\
    \hline  
\end{tabular}
\caption{Examples from Llama 2 7B Chat on QA, Summ, Data2txt, and FS. We color faithful sentences in blue and unfaithful ones in red. \modelname provides accurate detection of the locations of unfaithful output.}
\label{qualitative-study-selection}
\end{figure*}

\end{document}